%% file: main.tex
\documentclass[lettersize,journal]{IEEEtran}
\usepackage{amsmath,amsfonts}
\usepackage{algorithmic}
\usepackage{algorithm}
\usepackage{array}
\usepackage{textcomp}
\usepackage{stfloats}
\usepackage{url}
\usepackage{verbatim}
\usepackage{graphicx}
\usepackage{cite}
\usepackage{graphics} 
\usepackage{epsfig} 
\usepackage{mathptmx} 
\usepackage{times} 
\usepackage{amsmath} 
\usepackage{amssymb}  
\usepackage{algorithm}
\usepackage{algorithmic}
\usepackage{gensymb}
\usepackage{multirow}
\usepackage{subcaption}
\usepackage{cite}
\usepackage{outlines}
\usepackage{url}

\usepackage{hyperref} 

\makeatletter
\let\NAT@parse\undefined
\makeatother
\usepackage{hyperref}

\usepackage{color}

\title{\LARGE \bf
RoMe: Towards Large Scale Road Surface Reconstruction
\\via Mesh Representation
}

\author{Ruohong Mei$^{1*}$, Wei Sui$^{1}$, Jiaxin Zhang$^{1*}$, Xue Qin$^{2}$, Gang Wang$^{3}$, Tao Peng$^{1}$, Tao Chen$^{1}$ and Cong Yang$^{1\dag}$ }%

\begin{document}

\twocolumn[{
\renewcommand\twocolumn[1][]{#1}
\maketitle

\begin{center}   
    \centering
    \includegraphics[width=0.9\linewidth]{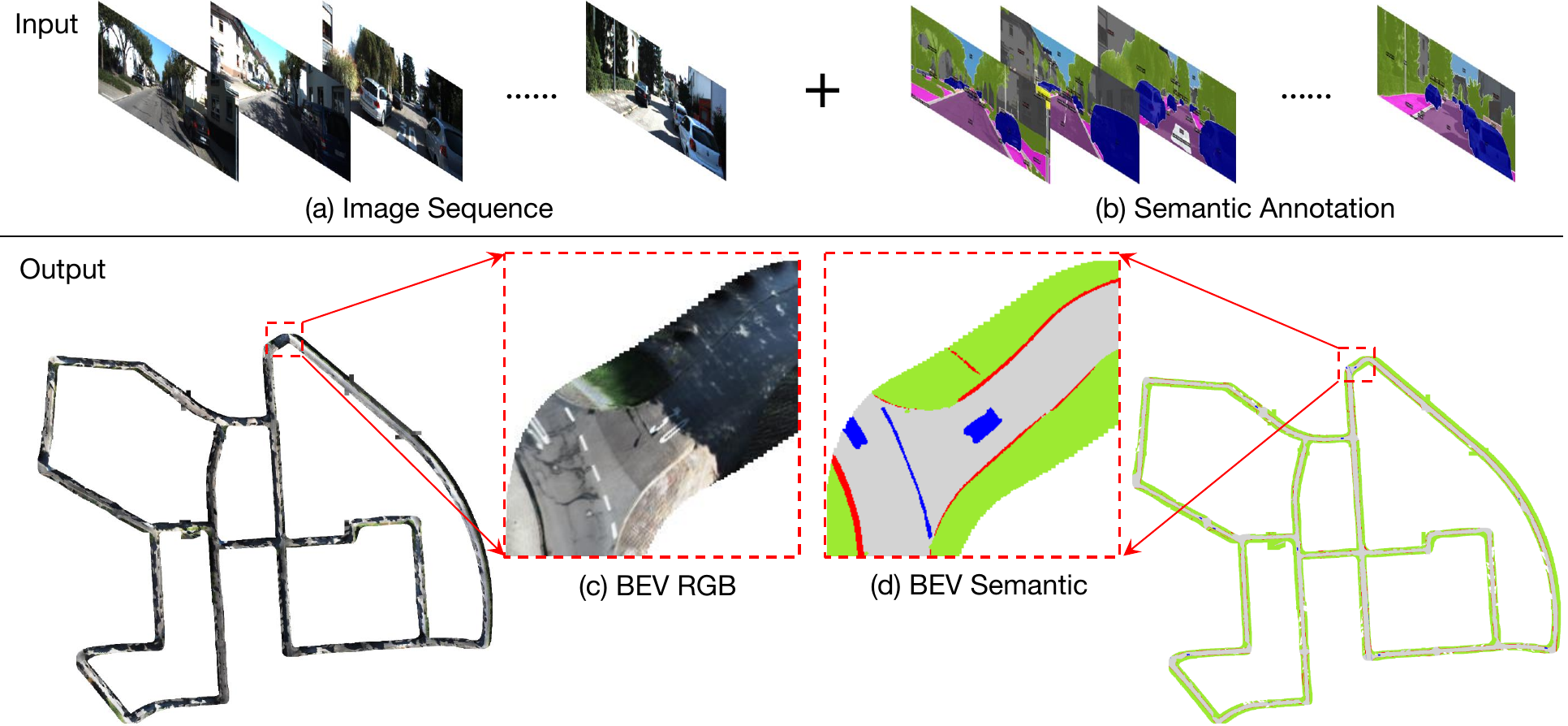}
    \captionof{figure}{\small Road surface reconstruction results (KITTI odometry sequence-00) using our proposed RoMe, covering an area of approximately $600\times600$ square meters. The first row displays the input image sequence with semantic annotations. The second row showcases the final results with close-up details highlighted in red rectangles: the reconstructed BEV RGB surface and its corresponding BEV semantics.
    }
   \label{fig:structure}
\end{center}%
}]

\let\thefootnote\relax\footnotetext{$^{1}$ R. Mei, J. Zhang, W. Sui, T. Peng, T. Chen and C. Yang are with BeeLab, School of Future Science and Engineering, Soochow University, Suzhou, China. * Equal contribution. $\dag$ Corresponding (cong.yang@suda.edu.cn).

$^{2}$ Xue Qin is with Harbin Institute of Technology, Harbin, China.

$^{3}$ Gang Wang is with Shandong University, Shandong, China.
}

\markboth{Journal of \LaTeX\ Class Files,~Vol.~14, No.~8, November~2023}%
{Shell \MakeLowercase{\textit{et al.}}: A Sample Article Using IEEEtran.cls for IEEE Journals}


\begin{abstract}
\input{abstract}
\end{abstract}

\begin{IEEEkeywords}
Road Surface Reconstruction, Multilayer Perception Network, Waypoint Sampling, Extrinsic Optimization.
\end{IEEEkeywords}

\IEEEpubidadjcol
\input{introduction}
\input{related}
\input{approach}
\input{experiments}
\input{conclusion}

\section*{Acknowledgments}
This work was supported in part by the Natural Science Foundation of the Jiangsu Higher Education Institutions of China (22KJB520008); in part by the Research Fund of Horizon Robotics (H230666); and in part by the Jiangsu Policy Guidance Program, International Science and Technology Cooperation, The Belt and Road Initiative Innovative Cooperation Projects (BZ2021016).

\bibliographystyle{IEEEtran}
\bibliography{ref}

\vspace{30 mm}
\begin{IEEEbiography}[{\includegraphics[width=1in,height=1.25in,clip,keepaspectratio]{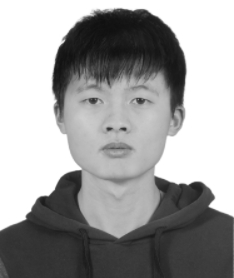}}]{Ruohong Mei}
is currently an Algorithm Engineer at Horizon Robotics in Beijing, China. He earned his B.S. Degree in Communication Engineering from Beijing University of Posts and Telecommunications in 2018, followed by an M.S. Degree in Information and Communication Engineering from Beijing University of Posts and Telecommunications in 2021. His primary research interests lie in 3D vision, and deep learning.
\end{IEEEbiography}

\begin{IEEEbiography}[{\includegraphics[width=1in,height=1.25in,clip,keepaspectratio]{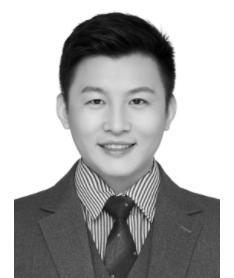}}]{Wei Sui}
is a senior engineer at Horizon Robotics, leading the 3D Vision Team, providing mapping, localization, calibration, and 4D labeling solutions. His research interests include SFM, SLAM, Nerf, 3D Perception, etc. Dr. Wei received his B.Eng and Ph.D. degrees from Beinghang University and NLPR (CASIA), Beijing, China, in 2011 and 2016 respectively. He led the computer vision team and successfully developed the 4D Labeling System and BEV perception for Super Drive on Journey 5. Dr. Wei Sui has published one research monograph and more than ten peer-reviewed papers in journals and conference proceedings, including elites like TIP, TVCG, ICRA, CVPR, etc. Dr. Wei received over 40 Chinese and 5 US patents.
\end{IEEEbiography}

\begin{IEEEbiography}[{\includegraphics[width=1in,height=1.25in,clip,keepaspectratio]{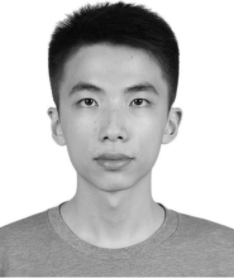}}]{Jiaxin Zhang}
is currently an Algorithm Engineer at Horizon Robotics in Beijing, China. He earned his B.S. degree in Applied Physics from the University of Science and Technology of China in 2018, followed by an M.S. degree in Electrical and Computer Engineering from Boston University in 2020. His primary research interests lie in SLAM, 3D vision, and deep learning.
\end{IEEEbiography}

\begin{IEEEbiography}[{\includegraphics[width=1in,height=1.25in,clip,keepaspectratio]{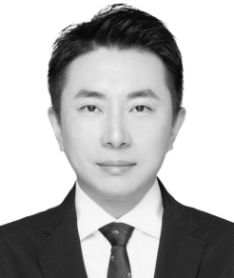}}]{Xue Qin} is a Senior Engineer affiliated with the Harbin Institute of Technology boasts an extensive academic background in the realm of computer science and technology. He commenced his academic journey with a Bachelor's degree in Computer Science from the University of Melbourne, followed by a Master's degree in Network Computing from Monash University. Presently, Qin is deepening his research endeavours as a PhD candidate in Computer Science and Technology at the Harbin Institute of Technology. Beyond his technical pursuits, he has also broadened his managerial acumen by completing an Executive Master of Business Administration from the prestigious Tsinghua University. His scholarly contributions predominantly revolve around artificial intelligence, computer vision, and pioneering anti-collision systems for autonomous vehicles. 
\end{IEEEbiography}

\begin{IEEEbiography}[{\includegraphics[width=1in,height=1.25in,clip,keepaspectratio]{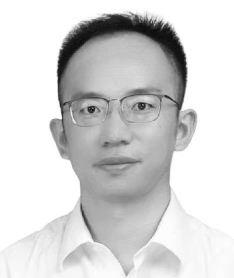}}]{Gang Wang}, 
master degree in Vehicle engineering from Wuhan University of Technology, PhD in intelligent manufacturing from Shandong University. In 2015, he began to work as a senior professional manager and senior engineer in automobile manufacturing enterprises. His main research direction is the application and industrialization of intelligent manufacturing technology, 3D vision technology and SLAM laser navigation technology in the direction of unmanned logistics. He has published many high-level papers and software Copyrights in the direction of industrial big data and manufacturing digital transformation.
\end{IEEEbiography}

\begin{IEEEbiography}[{\includegraphics[width=1in,height=1.25in,clip,keepaspectratio]{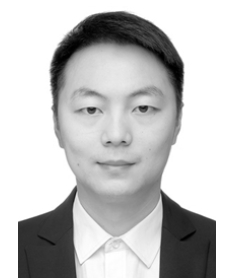}}]{Tao Peng}
is an Associate Professor in Soochow University, China, since 2022. Before that, Dr. Peng received his Ph.D. degree in the Department of Computer Science and Technology at Soochow University in 2019. From 2020 to 2022, he was a postdoctoral researcher in the Department of Health Technology and Informatics at Hong Kong Polytechnic University, and Department of Radiation Oncology at University of Texas Southwestern Medical Center, Dallas, USA, successively. During this period, he obtained the “Research Talent” award from Hong Kong government. He has published more than 40 peer-reviewed journal/conference papers, where the total impact factor (IF) of all the journal publications as the first author is IF > 98. He now serves as Guest Associate Editor of Medical Physics journal, a Co-Editor of Special Topic at Frontiers in Signal Processing journal, Program Committee of 20(th) PRICAI-2023 and iWOAR 2023 conference, and a reviewer of more than 20 high-quality journals/conferences. His main research interests include image processing, pattern recognition, machine learning, and their applications.
\end{IEEEbiography}

\begin{IEEEbiography}[{\includegraphics[width=1in,height=1.25in,clip,keepaspectratio]{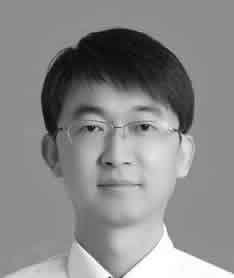}}]{Tao Chen}
received the B.Sc. degree in Mechanical Design, Manufacturing and Automation, M.Sc. degree in Mechatronic Engineering, and Ph.D. degree in Mechatronic Engineering from Harbin Institute of Technology, Harbin, China, in 2004, 2006, and 2010, respectively. He is a visiting scholar in National University of Singapore in 2018. He is currently an professor at School of Future Science and Engineering, Soochow University, Suzhou, China. His main research interests include MEMS, sensors, and actuators.
\end{IEEEbiography}

\begin{IEEEbiography}[{\includegraphics[width=1in,height=1.25in,clip,keepaspectratio]{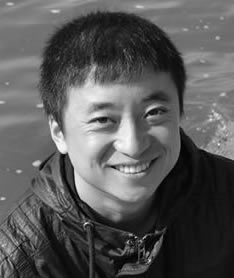}}]{Cong Yang}
is an Associate Professor at Soochow University since 2022. Before that, he was a Postdoc researcher at the MAGRIT team in INRIA (France). Later, he worked scientifically and led the computer vision and machine learning teams in Clobotics and Horizon Robotics. His main research interests are computer vision, pattern recognition, and their interdisciplinary applications. Cong earned his Ph.D. degree in computer vision and pattern recognition from the University of Siegen (Germany) in 2016.
\end{IEEEbiography}

\end{document}

%% file: abstract.tex
In autonomous driving applications, accurate and efficient road surface reconstruction is paramount. This paper introduces RoMe, a novel framework designed for the robust reconstruction of large-scale road surfaces. Leveraging a unique mesh representation, RoMe ensures that the reconstructed road surfaces are accurate and seamlessly aligned with semantics. To address challenges in computational efficiency, we propose a waypoint sampling strategy, enabling RoMe to reconstruct vast environments by focusing on sub-areas and subsequently merging them. Furthermore, we incorporate an extrinsic optimization module to enhance the robustness against inaccuracies in extrinsic calibration. Our extensive evaluations of both public datasets and wild data underscore RoMe's superiority in terms of speed, accuracy, and robustness. For instance, it costs only 2 GPU hours to recover a road surface of $600\times600$ square meters from thousands of images. Notably, RoMe's capability extends beyond mere reconstruction, offering significant value for auto-labeling tasks in autonomous driving applications. All related data and code are available at \href{https://github.com/DRosemei/RoMe}{GitHub}.


%% file: introduction.tex
\section{INTRODUCTION}
\label{sec:intro}
\IEEEpubidadjcol
\IEEEPARstart{I}{n} the realm of autonomous driving, bird-eye-view (BEV) perception has emerged as a pivotal tool, aligning seamlessly with tasks such as planning and control. This underscores the significance of large-scale road surface reconstruction, especially when it comes to training and validating BEV perception tasks. Broadly, road surface reconstruction methodologies can be bifurcated into two primary categories: traditional methods~\cite{schoenberger2016sfm,schoenberger2016mvs} and those anchored in neural radiance fields (NeRF)~\cite{mildenhall2020nerf,Rematas_2022_CVPR,li2022read,tancik2022block}.

Traditional Multi-View Stereo (MVS) approaches often yield dense point reconstructions. While these are adept for surfaces with distinct textures, they tend to falter, producing noisy and incomplete results for more uniform road surfaces. Furthermore, their computational demands escalate for expansive reconstructions. Conversely, recent advancements have witnessed the adoption of implicit representation-based methodologies for photorealistic reconstruction, utilizing a curated set of posed images~\cite{Rematas_2022_CVPR,li2022read,tancik2022block}. These leverage tools such as Multi-Layer Perceptions (MLP) to recreate intricate cityscapes. However, their extensive resource requirements often render them less feasible for large-scale applications.

\begin{figure*}[t!]
\centering
\includegraphics[width=1\linewidth]{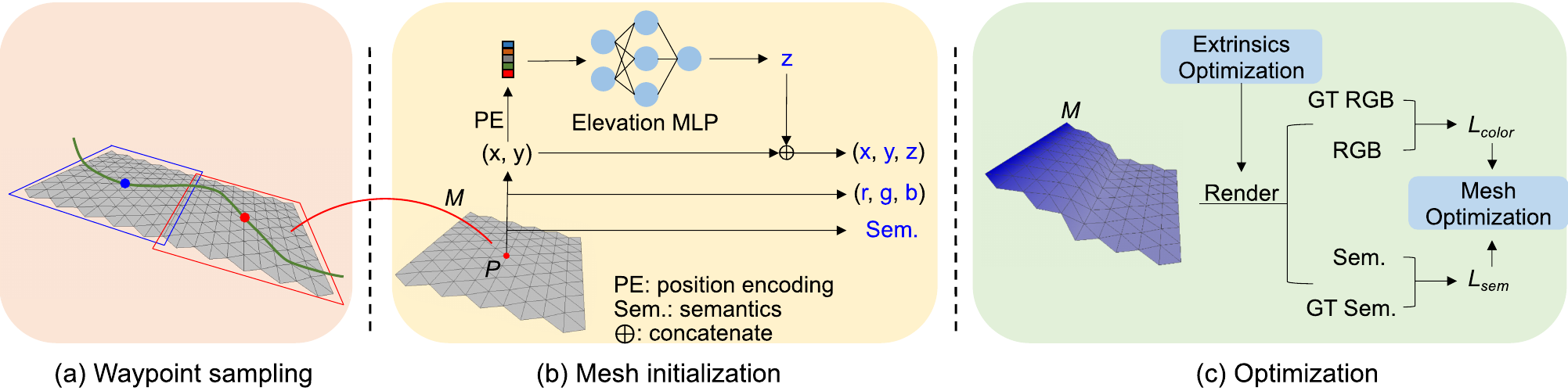}
\caption{\small Overview of RoMe. (a) Waypoint sampling: The green line depicts the camera's path. Red and blue boxes indicate neighboring subareas, with corresponding red and blue dots representing waypoint samples, aiding in faster training. (b) Mesh initialization: Upon initializing mesh $M$, vertices are assigned a position $(x, y, z)$, color $(r, g, b)$, and semantic attributes. The elevation $z$ of each vertex is fine-tuned using an elevation MLP network. (c) Optimization: The optimization targets, $L_{color}$ and $L_{sem}$, enable rendering mesh $M$ into RGB images with associated semantics. The parameters ($z$, $(r, g, b)$, and Sem., highlighted in blue in (b)) are collectively adjusted to produce the final road mesh $M$. Best viewed in color.}
\label{fig:mesh}
\end{figure*}

In real-world scenarios, 3D road surfaces often exhibit discontinuities, suggesting they can be delineated as smooth meshes with nuanced elevations. Motivated by this, we conceived \underline{RoMe} (\underline{Ro}ad \underline{Me}sh), a methodical approach for large-scale road surface reconstruction, reliant solely on images. As delineated in Fig.~\ref{fig:structure}, RoMe crafts a comprehensive 3D road mesh from a sequence of images, complemented by their semantic annotations. Each mesh vertex encapsulates details of elevation, color, and semantics. Fig.~\ref{fig:mesh} presents the general idea of RoMe: (1) Waypoint sampling: aims to expedite the reconstruction process via a divide-and-conquer strategy: iteratively reconstruct subareas (only a tiny portion of the current view) rather than the whole surface. Herein, the green trajectory epitomizes the camera's path, with the red and blue boxes demarcating adjacent subareas. (2) Mesh initialization: each vertex is encoded by position, color, and semantics. The elevation of each vertex is adeptly modeled via an MLP network. (3) Mesh optimization: focusing on color and semantics, facilitating the rendering of the mesh into RGB images with corresponding semantics. This intricate process ensures the joint optimization of parameters, culminating in the final reconstructed road mesh. To further boost the robustness of RoMe on cameras and environments, we also incorporate a mechanism to fine-tune the settings mentioned above during the reconstruction process.

In summary, our main contributions are: (1) a 2D implicit road surface representation method is introduced to achieve highly efficient reconstruction for road surfaces. (2) A waypoint sampling algorithm is proposed to reduce the memory and time costs: the whole system is able to reconstruct a road surface up to $600 \times 600$ square meters with only 2 GPU hours on an RTX-3090. Comprehensive experiments show that our proposed RoMe outperforms traditional methods and NeRF in road surface reconstruction tasks in terms of accuracy, efficiency, and robustness.

%% file: related.tex
\section{RELATED WORKS}
\label{sec:related}

Here, we briefly glanced through several existing multi-view stereo strategies, followed by a review of surface reconstruction methods. For a more detailed treatment of this topic in general, the recent compilation by~\cite{musialski2013survey, wang2021survey, gao2023nerf} offer a sufficiently good review.

\subsection{Multi-View Stereo}
3D reconstruction is a process of deducing the three-dimensional structure of an object or scene using multiple images captured from varied camera positions. This domain has witnessed significant advancements over the years~\cite{ozyecsil2017survey}. While effective in specific contexts, traditional Multi-View Stereo (MVS) methods often hinge on extracting and matching feature points. The performance is limited in texture-less scenes (e.g., road surface) where feature points are sparse and unevenly distributed~\cite{schoenberger2016mvs,schoenberger2016sfm}. Novel view synthesis, which produces photo-realistic images from previously unseen perspectives, shares a close affinity with MVS techniques. While some methods like~\cite{fan2018road, yu20073d, brunken2020road} are tailored for road surface reconstruction, their scope is limited to smaller areas, making them unsuitable for expansive scenarios. Large-scale MVS methods, applicable even at city levels, have been proposed~\cite{agarwal2011building}. These typically involve extracting points from images, constructing sparse 3D points, and subsequently generating meshes. However, they primarily target building structures, often overlooking texture-less surfaces like roads.

Our RoMe approach stands distinct, capable of reconstructing entire road surfaces irrespective of texture variations. It excels in reconstructing expansive road surfaces while preserving essential features such as textures, semantics, and elevations.

\subsection{Surface Reconstruction}
Existing MVS methods are not computationally efficient for road surface reconstruction since they model whole scenes through dense point clouds. In practice, existing road surface reconstruction techniques can be broadly categorized into explicit and implicit methods. For the first one, Tong et al.~\cite{qin2021light} introduced a system that leverages cameras to construct large-scale semantic maps. However, these methods heavily depend on Inverse Perspective Mapping (IPM) and may overlook elevation variations on road surfaces. Rendering-based techniques~\cite{levoy1996light, waechter2014let} employ mesh representations with view-dependent appearances. 

For the second one, implicit surface reconstruction has gained momentum with the advent of NeRF~\cite{mildenhall2020nerf}, which utilizes implicit representation and voxel rendering to achieve impressive Novel View Synthesis (NVS) results. Large-scale NeRF techniques aim to capture intricate details of city blocks or driving scenes. However, they often necessitate additional data acquisition tools, such as LiDAR and images from varied angles~\cite{Rematas_2022_CVPR, li2022read, tancik2022block}. In contrast, RoMe operates efficiently with a few vehicle-mounted cameras, making it compatible with platforms like nuScenes~\cite{caesar2020nuscenes} and KITTI~\cite{Geiger2013IJRR}. Besides, our proposed waypoint sampling approach can dramatically improve the reconstruction efficiency via a divide-and-conquer strategy, which is more friendly for parallel computing. 

\begin{figure}[t!]
   \begin{center}
      \includegraphics[width=1.0\linewidth]{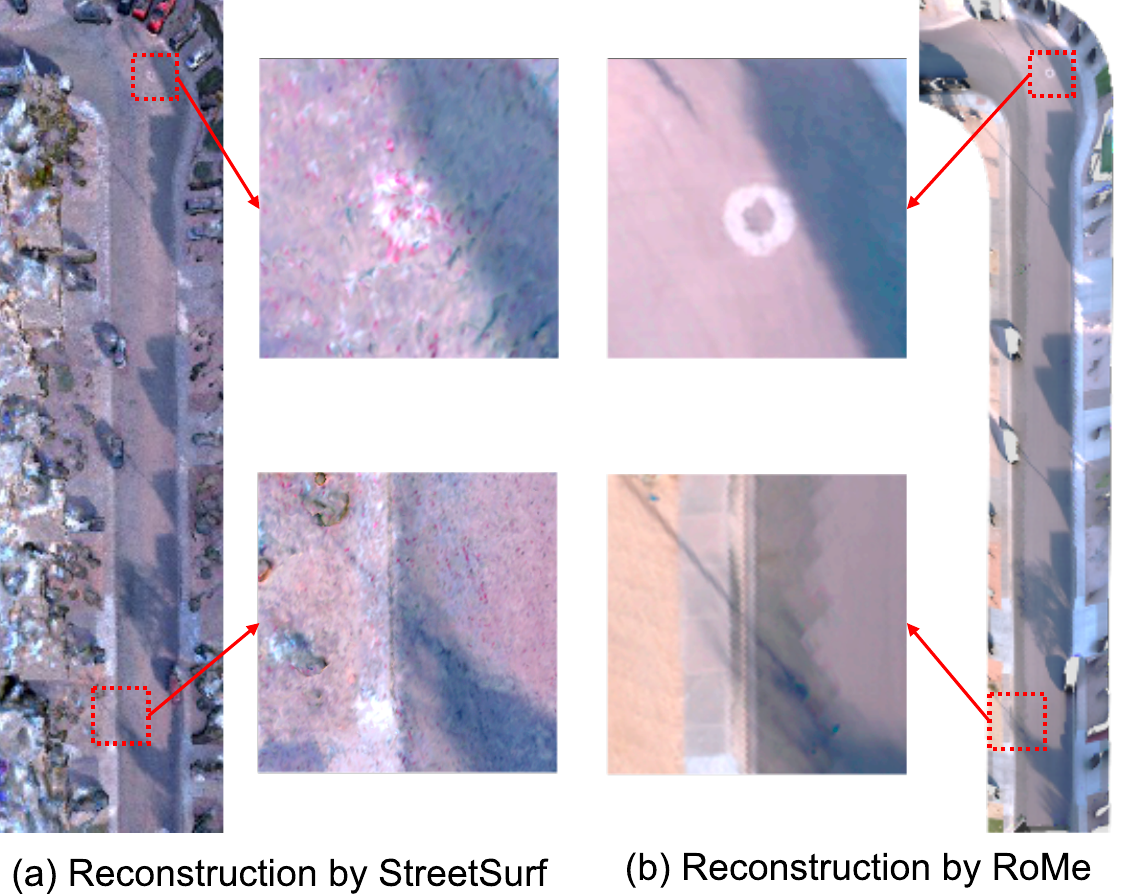}
   \end{center}
   \caption{\small Street reconstruction by StreetSurf~\cite{guo2023streetsurf} and RoMe.}
   \label{fig:streetsurf}
   \vspace{-1em}
\end{figure}
Noted that some pure-vision NeRF-based methods specifically target road surface reconstruction. For instance, Xie et al.~\cite{xie2023mvmap} employs a voxelized neural radiance field to refine High-Definition Maps (HD-Maps). Wang et al.~\cite{wang2023planerf} introduces a plane regularization technique based on Singular Value Decomposition (SVD) to optimize NeRF's 3D structure. However, these methods are sensitive to camera pose variations~\cite{bian2023nope}. RoMe, on the other hand, represents the road surface as a 3D mesh, optimizing it using multiple image supervisions, ensuring consistency and resilience to camera pose fluctuations. Though Guo et al.~\cite{guo2023streetsurf} segments the unbounded space into distinct sections (see Fig.~\ref{fig:streetsurf} (a)), its road surface mesh is blurry and lacks semantics and textures. In contrast, our mesh is smooth, watertight, texture-rich, and properly preserves the original semantics (see Fig.~\ref{fig:streetsurf} (b)).

%% file: approach.tex
\section{APPROACHES}
\label{sec:approach}
RoMe aims to reconstruct road surface textures and semantics using a sequence of images. As illustrated in Fig.~\ref{fig:mesh}, RoMe comprises three primary components: Waypoint Sampling, Mesh Initialization, and Optimization. For clarity in terminology, commonly used terms and expressions are defined:
\begin{itemize}
    \item Ego: self-vehicle, usually same as the mounting position of Inertial Measurement Unit (IMU)/Global Navigation Satellite Systems (GNSS).
    \item Ego pose: self-vehicle transforms in world coordinate.
    \item Camera pose: camera transforms in world coordinate.
    \item Elevation: road surface elevation in world coordinate.
    \item Waypoints: points that divide road surface to sub-areas for faster reconstruction.
\end{itemize}


\subsection{Mesh Initialization}
\label{s:approach:init}
Mesh initialization relies on camera poses estimated using ORB-SLAM2~\cite{campos2021orb} (or COLMAP~\cite{schoenberger2016sfm}). ORB-SLAM2 is a real-time SLAM library for monocular, stereo, and RGB-D cameras that computes camera poses and a sparse 3D reconstruction. For instance, we use stereo cameras in KITTI for restoring camera poses. Then, the semantic segmentation method Mask2Former~\cite{cheng2022masked} is employed to generate semantics, including roads, curbs, sign lanes, vehicles, etc. Particularly, Mask2Former has robust and state-of-the-art performance on driving datasets like Cityscapes~\cite{Cordts2016Cityscapes} and Mapillary Vistas~\cite{neuhold2017mapillary}. These semantics are also used to mask out dynamic objects like vehicles and pedestrians, which could disrupt the consistency of the overall road structure.

We draw inspiration from~\cite{guo2023streetsurf} to achieve a more accurate mesh initialization. Specifically, we extend the ego poses horizontally to obtain semi-dense points. These points are then lowered by approximately equivalent to the ego height. This process yields points that are closer to the road surface. Pretraining the elevation MLP with these points aids in restoring elevation, especially in the areas with steep slopes. In Fig.~\ref{fig:mesh}, the initialized flat mesh, denoted by $M$, consists of equilateral triangles. Each face has three vertices, each vertex $P$ possessing attributes including location $(x, y, z)$, color $(r, g, b)$, and semantics. The position encoding is applied to $(x, y)$, subsequently feeding them into the elevation $\mathrm{MLP}$ to predict elevation $z$ as per Eq.~\ref{elevation_mlp}. The rationale behind using $\mathrm{MLP}(\cdot)$ is to control the smoothness of the road surface by adjusting the frequency of $\mathrm{PE}$.
\begin{equation}
\small
z = \mathrm{MLP}(\mathrm{PE}(x, y))
\label{elevation_mlp}
\end{equation}

\subsection{Waypoint Sampling}
\label{s:approach:ws}
\begin{figure}[t!]
   \centering
   \includegraphics[width=0.7\linewidth]{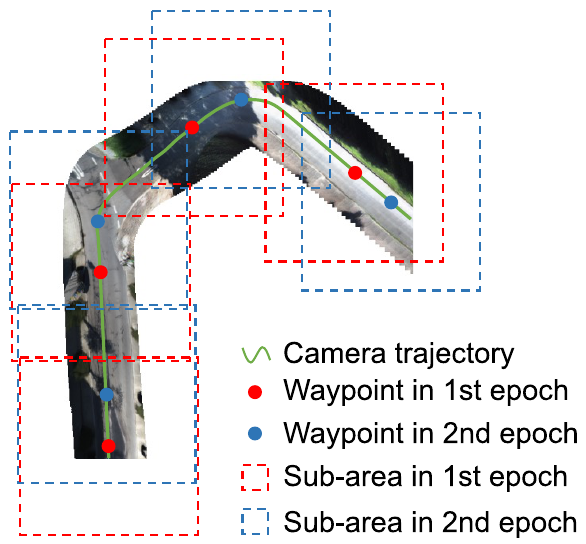}
   \caption{\small Illustration of waypoint sampling. The camera trajectory is represented by the green line. Distinct colored dots and their associated boxes indicate sampled waypoints and their corresponding sub-areas across various epochs.}
   \label{fig:waypoint}
\end{figure}
\begin{figure}[t!]
\renewcommand{\figurename}{Figure}
  \centering
  \includegraphics[width=1\linewidth]{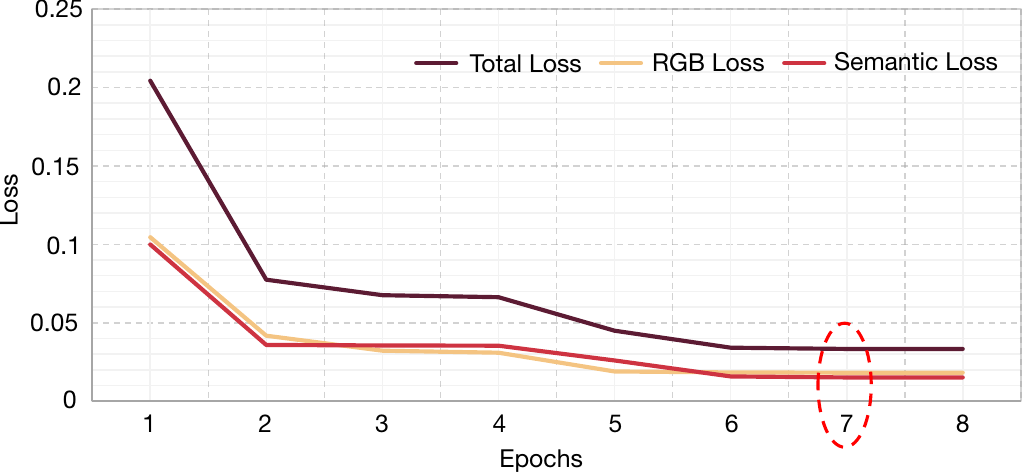}
  \caption{\small Epochs and losses in our preliminary experiments.}
  \label{fig:rome_alg}
  \vspace{-1.5em}
\end{figure}

To expedite the reconstruction of large areas (e.g., $600 \times 600$ square meters), we introduce a novel waypoint sampling approach to improve the efficiency of mesh initialization in Section~\ref{s:approach:init}. As presented in Fig.~\ref{fig:waypoint}, the core principle is divide-and-conquer. In other words, instead of reconstructing the entire road surface in one go, RoMe divides the vast area into smaller, manageable sub-areas centered around waypoints. Each of these sub-areas is then reconstructed individually. Once all sub-areas are processed, they are seamlessly merged to form the complete road surface reconstruction. It enhances computational efficiency and ensures detailed representation across the entire area.
\begin{figure}[t!]
  \centering
  \includegraphics[width=1.0\linewidth]{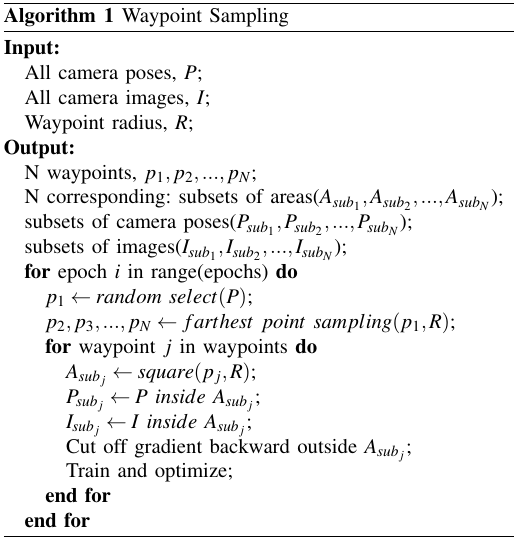}
  \vspace{-3em}
\end{figure}
As detailed in Algorithm 1, camera pose positions are treated as a set of point clouds $P$. The first waypoint is randomly selected from $P$. Given the desired sampling radius $R$, the farthest point sampling algorithm selects waypoints ($ p_1,p_2,...,p_N$). Subsequent steps involve gathering all camera poses $P_{sub_j}$ and images $I_{sub_j}$ within the radius for each waypoint $p_j$ and traversing each sub-area $A_{sub_j}$ for optimization. This process is iteratively applied until all sub-areas are adequately covered, resulting in the entire road surface being updated. In practice, the initial waypoint is randomly selected in each training epoch to ensure consistency at the boundaries between different sub-areas. We detail the stopping conditions for two loops in Algorithm 1: 

\begin{outline}
 \1 \textbf{The inner loop} is for waypoint sampling. The index $j$ iterates over each waypoint. The farthest point sampling determines the number of the waypoints. There are around five waypoints for a typical $300\times300$ reconstruction area. The details of farthest point sampling are: (1) Gather all ego vehicle poses as point clouds with position $x,y,z$. (2) Randomly select a point from the point cloud for starting. (3) Given preset parameter radius $R$, draw a circle as the start area. All the points within $R$ are marked selected. (4) Pick the farthest point from the unselected points and draw a circle with $R$. (5) Repeat (2)-(4) until all points in the point clouds are selected.
 \1 \textbf{The outer loop} is for training with multiple epochs. Our preliminary experiments (see Fig.~\ref{fig:rome_alg}) show that seven epochs are good enough to balance the point coverage and computing requirements.
 \1 \textbf{Stopping Condition}. The stop condition for the inner loop is the number of waypoints (5 waypoints). The stop condition for the outer loop is the number of epochs (7 epochs).
\end{outline}
\subsection{Optimization}
\label{s:approach:opt}
Our optimization strategy has twofold: (1) Extrinsic optimization to improve the robustness of RoMe on various camera settings and (2) Mesh optimization during the training process on color and semantics.

\subsubsection{Extrinsic Optimization}
\label{s:approach:opt:extrinsic}
In the context of camera calibration, extrinsic refers to the parameters that define the position and orientation of the camera in a world coordinate system. They capture the relationship between the camera's local coordinate system and a global, fixed coordinate system. Accurate camera extrinsic is not always guaranteed. For instance, we observed that the extrinsic among nuScenes cameras are not always ideal in some scenes. Ego poses pertain to the position and orientation of the autonomous vehicle (or ego vehicle) within its environment. It provides a reference frame from which other objects and landmarks can be localized. In our approach, we decouple camera poses into vehicle ego poses and camera extrinsic. Camera extrinsic describes the transformation between the vehicle coordinate system (often called the ego coordinate system) and the camera coordinate system. This transformation is crucial for aligning the visual data captured by the camera and other sensors on the vehicle.

In RoMe, camera extrinsic is expressed as a transform matrix $T=[R|t] \in$ SE(3), where $R \in$ SO(3) and  $t \in {\mathbb{R}\\}^3$  denote rotation and translation, respectively. Translation $t$ can be easily optimized because it is defined in Euclidean space. Rotation $R$ is expressed as the axis-angle: $\phi := \alpha \omega$, $\phi \in {\mathbb{R}\\}^3$, where $\alpha$ is a rotation angle and $\omega$ is a normalized rotation axis. It can be converted to $R$ by Rodrigues' formula:
\begin{equation} \label{rodrigues}
   R = I + \frac{sin(\alpha)}{\alpha} \phi^{\land} + 
      \frac{1-cos(\alpha)}{\alpha^2} (\phi^{\land})^2
\end{equation}
in which skew operator $(\cdot)^{\land}$ converts a vector $\phi$ to a skew 
matrix:
\begin{equation} \label{skew}
   \phi^{\land} = {\begin{pmatrix} 
      \phi_0 
    \\\phi_1
    \\\phi_2
   \end{pmatrix}}^{\land} = \begin{pmatrix}
     0 & -\phi_2 & \phi_1\\
     \phi_2& 0 & -\phi_0\\
     -\phi_1 & \phi_0 & 0
   \end{pmatrix}
\end{equation}

In practice, we optimize relative camera extrinsic compared with calibrated extrinsic by $\alpha$, $\phi$, and translation $t$ for faster and easier convergence.

\subsubsection{Mesh Optimization}
\label{s:approach:opt:mesh}
To derive training supervision, we first input the source mesh \(M\) into the differentiable renderer in~\cite{ravi2020accelerating}. Specifically, as shown in Eq.~\ref{eq1}, we apply a MeshRenderer function to \(M\) to obtain rendering results of image views from the \(j\)-th camera pose \({\pi}_j\):

\begin{equation}
\small
[C_j, S_j, D_j, Mask_j] = MeshRender({\pi}_j, M)
\label{eq1}
\end{equation}
where \(C_j\), \(S_j\), and \(D_j\) represent the \(j\)-th rendered RGB, semantic, and depth images, respectively. \(Mask_j\) is the corresponding silhouette image indicating the area of supervision. \(N\) is the maximum number of source images and corresponding poses. \(j=1,\cdots, N\). \(D_j\) could be supervised if sparse or dense depth is provided. The specific procedure is as follows:
\begin{outline}
 \1 Transform the mesh vertices coordinates into $j_{th}$ frame:
 \begin{equation}
 \small
 V_j = T_j*V_M
 \label{eq2}
 \end{equation}
 where $V_M$ is the vertices of the Mesh $M$ in world coordinate, $V_j$ is the vertices of Mesh in $j_{th}$ frame coordinate and $T_j$ is the transform matrix.
 \1 Compute fragments for each frame given transformed mesh $M_j$. The fragments consist of the components:
   \2 Pixel-to-faces are tensors, giving the indices of the nearest faces at each pixel, sorted in ascending z-order.

   \2 z-buffers are tensors, giving the NDC z-coordinates of the nearest faces at each pixel, sorted in ascending z-order.

   \2 Barycentric coordinates are tensors, giving the barycentric coordinates in NDC units of the nearest faces at each pixel, sorted in ascending z-order.

   \2 Distances are tensors, giving the signed Euclidean distance (in NDC units) in the x/y plane of each point closest to the pixel.
 \1 Render the fragments into images and depths. The images are rendered with hard channel blending, which is the naive blending of top K faces to return a new image. The depths are derived directly from the z-buffer.
\end{outline}

Building on Eq.~\ref{eq1}, we define the color (aka. RGB) loss \(L_{color}\) and the semantic loss \(L_{sem}\) for training RGB images and semantics, respectively:
\begin{equation}
\small
L_{color} = \frac{1}{N*sum(Mask_j)}\sum_{j=1}^{N} Mask_j * |C_j - \bar{C}_j|
\label{eq2}
\end{equation}
\begin{equation} 
\small
L_{sem} = \frac{1}{N*sum(Mask_j)}\sum_{j=1}^{N} Mask_j * CE(S_j,\bar{S}_j)
\label{eq3}
\end{equation}
where \(\bar{C}_N\) and \(\bar{S}_N\) denote the ground truth of RGB images and semantics, respectively. \(CE(\cdot)\) refers to the cross-entropy loss. During training, each vertex is optimized by multiple images from different views. Once all of them (from thousands to millions depending on mesh resolution) are properly optimized, the final mesh (with elevation, colors, and semantics) is obtained to represent the whole road surface.

\subsection{Implementation}
\label{s:approach:implementation}
RoMe initializes a road surface mesh based on~\cite{ravi2020accelerating}, utilizing RGBs, semantics, and elevation. Adam optimizer~\cite{2014Adam} is used during the training. We set the learning rates for BEV RGBs, semantics, and elevations at 0.1, 0.1, and 0.001, respectively. Typically, running the model for 7 epochs, with a halving of the learning rate at the 2nd and 4th epochs, yielded satisfactory results for most scenes. We set the BEV resolution at 0.1 meters/pixel. The Elevation MLP, an 8-layer network with a width of 128, is adapted from~\cite{wang2021nerf}. Table~\ref{table:training_settings} details the essential training parameters. Our experiments in Section~\ref{s:exp} were applied on a Linux server with a single RTX-3090 GPU.

\begin{table}[t!]
\small
\centering
\renewcommand{\tablename}{TABLE}
   \caption{\small Training details in our experiments.}
   \begin{tabular}{l|l}
   \hline
    Training settings & Values \\
   \hline
    \multirow{4}{*}{Learnable parameters}
     & {BEV RGB} \\
     & {BEV semantics} \\
     & {BEV elevations (MLP)} \\
     & {Extrinsics} \\
    \hline
    \multirow{2}{*}{Target functions}
    & {L1 loss for BEV RGB} \\
    & {Cross-entropy loss for BEV semantics} \\
    \hline
    Epochs & 7 \\
    \hline
    Batch size & 4 \\
    \hline
    Position encoding & 5 \\
    \hline
    Optimizer & Adam optimizer \\
    \hline
    \multirow{4}{*}{Learning rate}
     & {BEV RGB: 0.1} \\
     & {BEV semantics: 0.1} \\
     & {BEV elevations: 0.001} \\
     & {BEV extrinsics: 0.002} \\
    \hline
    Learning rate milestones & \{1, 4\} \\
    \hline
    Learning rate gamma & 0.1 \\
   \hline
   \end{tabular}
   \label{table:training_settings}
\end{table}

%% file: experiments.tex
\section{EXPERIMENTS}
\label{s:exp}
\begin{figure}[t!]
   \centering
   \includegraphics[width=\linewidth]{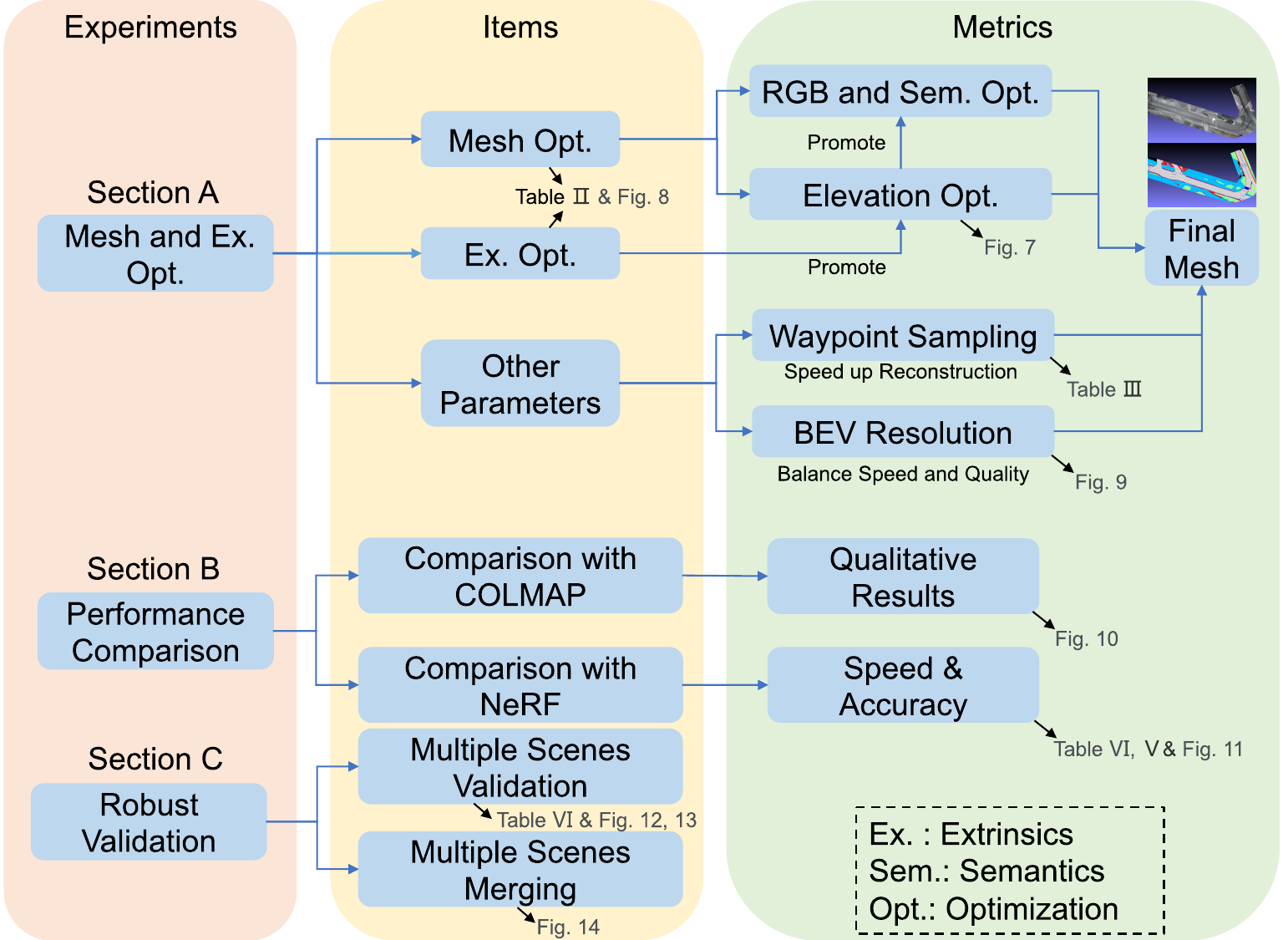}
   \caption{\small Workflow of our experiments.}
   \vspace{-1.5em}
   \label{fig:experiments}
\end{figure}

In this section, we first introduce the experimental setting, including datasets and metrics. After that, as presented in Fig.~\ref{fig:experiments}, we conduct our experiments with three major parts: 
\begin{itemize}
    \item In Section~\ref{s:exp:opt}, we subdivide experiments into mesh optimization, extrinsic optimization, and other parameters that affect final reconstruction results. Mainly, mesh optimization can be divided into RGB and semantics (with learnable parameters) and elevation (with MLP networks) optimization. Extrinsic optimization promotes elevation optimization, followed by RGB and semantics optimization to get final finer reconstruction results. Waypoint sampling speeds up the reconstruction, and BEV resolution can balance the speed and quality.
    \item In Section~\ref{s:exp:com}, we compare RoMe with COLMAP on quality and vanilla NeRF on novel view synthesis tasks in a single scene. Additionally, we compare RoMe with F2-NeRF~\cite{wang2023f2} on speed and accuracy, showcasing the superiority of RoMe in road surface reconstruction tasks.
    \item In Section~\ref{s:exp:robust}, we conduct experiments on 100 scenes chosen from nuScenes for multiple-scene validation to show RoMe's robustness and efficiency in merging multiple scenes to reconstruct larger areas.
\end{itemize}
\begin{figure}[t!]
   \centering
   \includegraphics[width=0.9\linewidth]{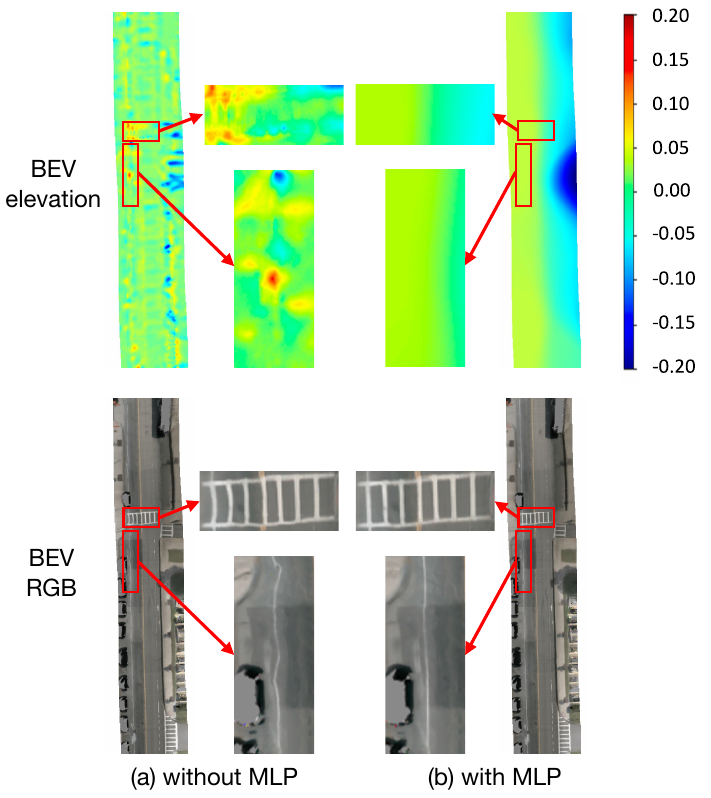}
   \caption{\small Ablation study on BEV elevation learning methods. The colourmap jet displays BEV elevation ranging from -0.2 meters to 0.2 meters. Utilizing MLP results in smoother elevation, enhancing reconstruction quality (highlighted in red boxes).}
   \label{fig:mlp}
   \vspace{-2em}
\end{figure}
\noindent{\bf Datasets:} We conduct our experiments on two renowned driving datasets: nuScenes~\cite{caesar2020nuscenes} and KITTI~\cite{Geiger2013IJRR}. The nuScenes dataset encompasses 1000 scenes, each being a 20-second video clip annotated at a frequency of 2 Hz. This dataset utilizes a camera rig with six cameras, providing a comprehensive 360-degree field of view. On the other hand, the KITTI odometry benchmark comprises 22 sequences, split into 11 training sequences (00-10) and 11 test sequences (11-21). For our experiments, we exclusively use monocular images from KITTI's left RGB camera. For semantic annotations, we employ the predictions from Mask2Former~\cite{cheng2022masked} with a Swin-L~\cite{liu2021swin} backbone since it has a state-of-the-art performance on primary semantic segmentation datasets, such as the Mapillary Vistas~\cite{neuhold2017mapillary}. For the nuScenes, we randomly selected 100 scenes, and for the KITTI, we chose sequence (00) for evaluation.

\noindent{\bf Metrics:} We assess the performance of all methods using standard NVS metrics: PSNR for image quality and mIoU for semantic segmentation accuracy. Following the convention of StreetSurf~\cite{guo2023streetsurf}, we adopt CD loss to evaluate the reconstruction geometry quality for full comparison besides depth RMSE metrics. The CD loss is an evaluation metric between two point clouds. It takes the distance of each point into account. In particular, CD finds the nearest point in the other point set and sums the square of distance up. We involve converting depth rendered from meshes and LiDAR depth into world coordinates to obtain point clouds:
\begin{equation} \label{CD loss}
   CD(\hat{G}, G) = \frac{1}{\hat{G}} \sum_{x\in \hat{G}}\min_{y\in G}\left \| x-y \right \| _{2}^{2} + \frac{1}{G} \sum_{y\in G}\min_{x\in \hat{G}}\left \| y-x \right \| _{2}^{2}
\end{equation}
where $\hat{G}$ and $G$ denote point clouds rendered from meshes and LiDAR depth, respectively. Besides, $x$ and $y$ denote 3D points in corresponding point clouds. We restrict our evaluation to points with semantic classes that are expected to be flat. To filter out outlier observations in LiDAR points, we compute the closest $97\%$ points in Chamfer distance, following the approach in~\cite{guo2023streetsurf}. Additionally, we utilize the RMSE metric to gauge the discrepancy between LiDAR depth and depth rendered from meshes.

\subsection{Mesh and Extrinsic Optimization}
\label{s:exp:opt}
\subsubsection{Mesh Optimization}
Mesh optimization is composed of RGB, semantics, and elevation optimizations. RGB and semantics optimization use the presentation of learnable parameters due to their high-frequency details. In terms of elevation optimization, it should be smooth in most cases, so we initiate our experiments by exploring two methods for BEV elevation representation. The first one treats BEV elevation as independent optimizable parameters (aka. ``with MLP"), similar to RGB and semantics. The second one utilizes an MLP representation (aka. ``with MLP"). As depicted in Fig.~\ref{fig:mlp}, BEV elevations without using MLP representation are not as smooth as the others. As a result, BEV RGBs are distorted, as shown in red boxes. Through our experiments, we observe that setting the position encoding frequency to 5 is adequate for most scenes.

\begin{table}[t!]
\small
\centering
   \setlength{\tabcolsep}{10pt}
   \caption{\small Ablation study on optimizing elevation and extrinsic. The best reconstruction results for both textures and semantics are achieved when both are optimized (highlighted in bold).}
   \begin{tabular}{c c|c c}
   \hline
    Opt. elevation & Opt. extrinsic & PSNR$\uparrow $ & mIoU ($\%$)$\uparrow $ \\
   \hline
    $\times$ & $\times$ & 25.5 & 71.4 \\
    $\surd$ & $\times$ & 25.9 & 73.9 \\
    $\times$  & $\surd$ & 26.0 & 79.0 \\
    $\surd$ & $\surd$ & \textbf{26.7} & \textbf{83.0} \\
   \hline
   \end{tabular}
   \label{table:ablation_study}
\end{table}
\begin{table}[t!]
\small
\centering
   \setlength{\tabcolsep}{4pt}
   \caption{\small Waypoint sampling efficiency. Utilizing waypoint sampling, we achieve a 2x speed-up and reduced GPU resource consumption without compromising the results.}
   \begin{tabular}{c|c c c c}
   \hline
   Waypoint Sampling & Time(min) & GPU(GB) & PSNR$\uparrow $ & mIoU($\%$)$\uparrow $ \\
   \hline
   $\times$ & 7 & 15 & 23.4 & 86.7 \\
   $\surd$ & 3.5 & 11 & 23.4 & 86.7 \\
   \hline
   \end{tabular}
   \label{table:waypoint_sampling}
   \vspace{-1em}
\end{table}
\begin{figure*}[t!]
  \centering
  \includegraphics[width=0.9\linewidth]{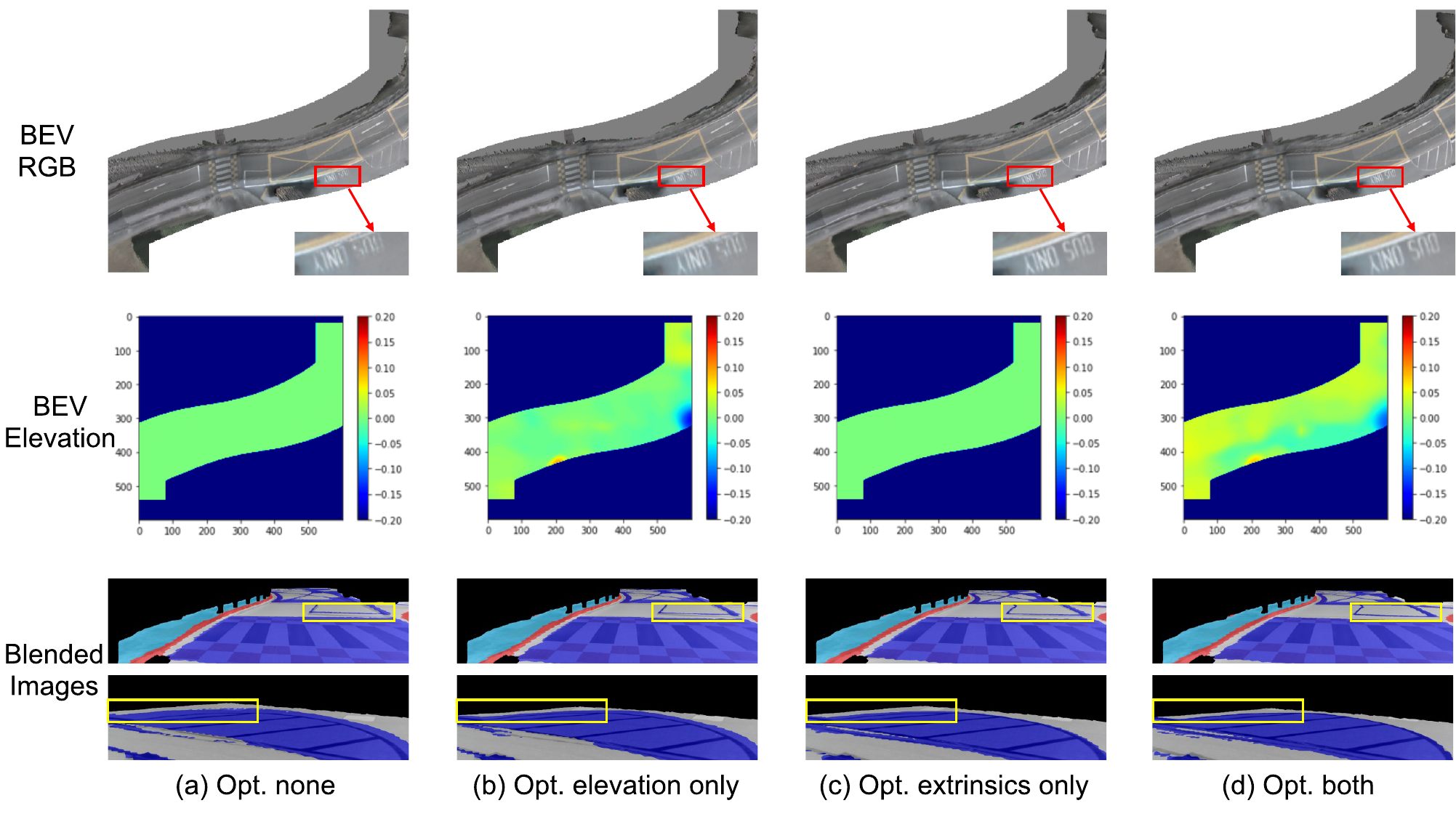}
  \caption{\small Ablation study results for elevation and extrinsic. The top panel shows BEV RGB, the middle displays BEV Elevation, and the bottom presents Blended Images. Comprehensive results can be found in Table~\ref{table:ablation_study}. Enhanced elevation restoration and extrinsic optimization lead to improved alignment of RGB and semantics.}
  \label{fig:ablation_study}
  \vspace{-1em}
\end{figure*}
\subsubsection{Extrinsic Optimization}
RoMe can restore road surface elevation and refine camera extrinsic, leading to a more precise reconstruction. For verification, we select a short clip from the $scene$-$0865$ of the nuScenes dataset. As detailed in Table~\ref{table:ablation_study}, implementing either elevation restoration or extrinsic optimization enhances the reconstruction results. Moreover, we observe that segmentation results are more sensitive to extrinsic inaccuracies. Some results are illustrated in Fig.~\ref{fig:ablation_study} for a more visual understanding. The top row displays the BEV RGB. The results appear blurry without applying elevation estimation or extrinsic optimization, especially in areas highlighted by red boxes. The middle row visualizes the BEV elevation (in meters). Notably, the BEV elevation in Fig.~\ref{fig:ablation_study} (d) exhibits more fluctuations than the others. The bottom row showcases blended images of the rendered semantics and the original images. A closer look at the yellow boxes reveals that without optimizing elevation and extrinsic, the rendered semantics do not align accurately with the source images.
\begin{figure}[t!]
   \centering
   \includegraphics[width=1.0\linewidth]{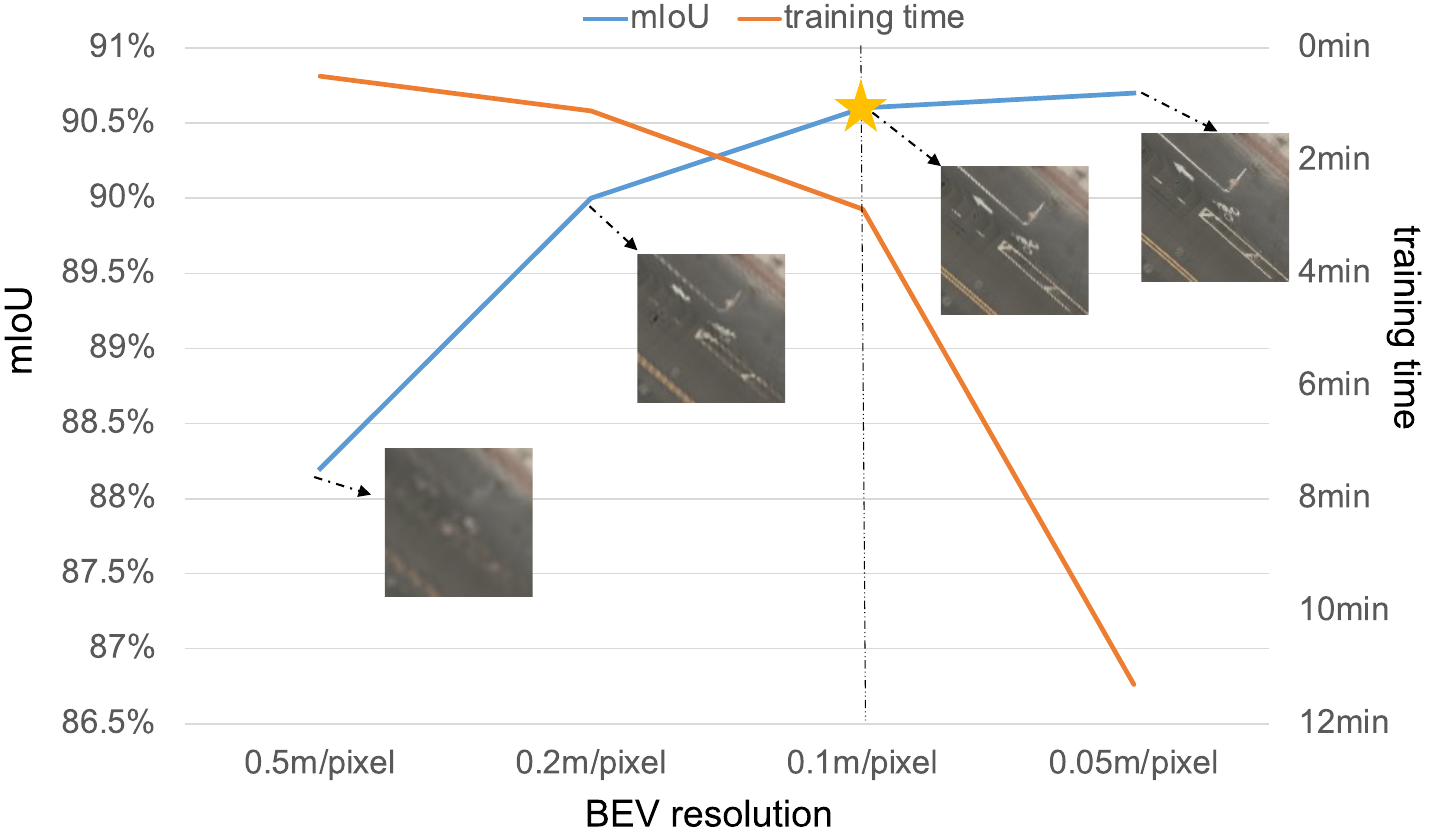}
   \caption{\small Ablation study on BEV resolution. A resolution of 0.1m/pixel achieves realistic reconstruction with improved training speed.}
   \label{fig:rome_bevres}
   \vspace{-1.8em}
\end{figure}
\begin{figure}[t!]
   \centering
   \includegraphics[width=0.9\linewidth]{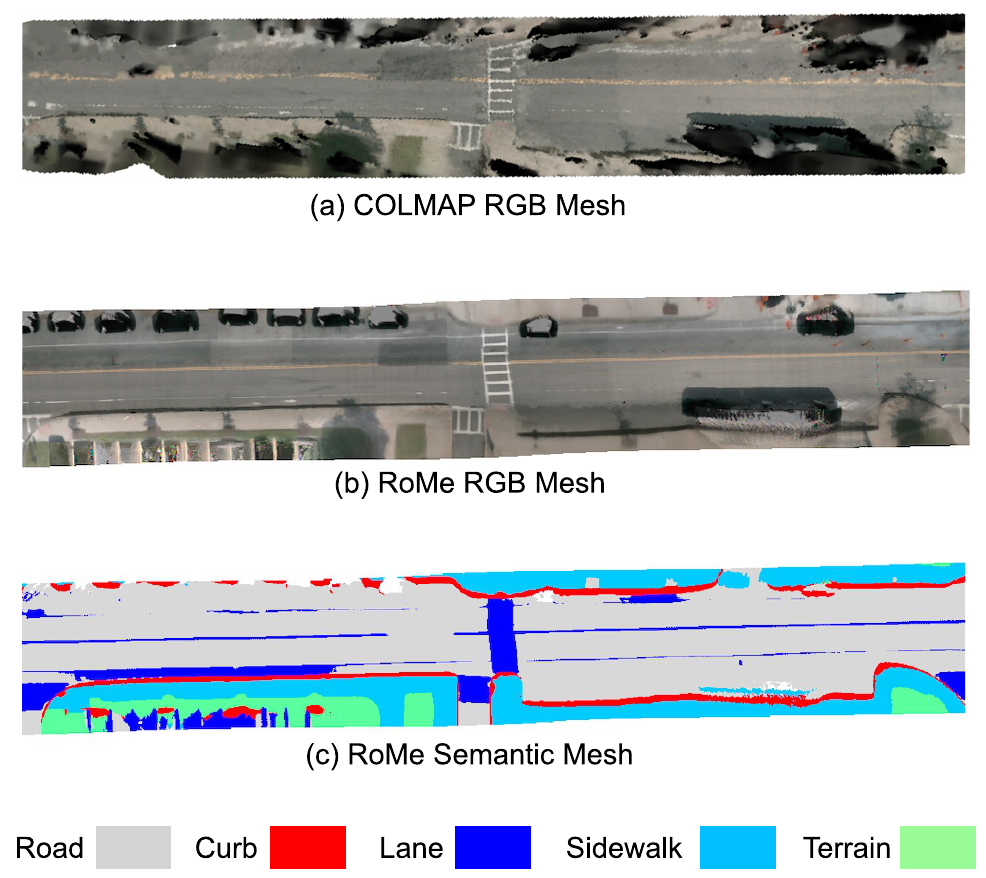}
   \caption{\small Comparison with COLMAP. While COLMAP may produce holes in the presence of moving objects, RoME remains robust, reconstructing from unobstructed frames. Additionally, RoME simultaneously reconstructs semantics.}
   \label{fig:colmap}
   \vspace{-1.8em}
\end{figure}
\subsubsection{Other Parameters}
To assess the efficiency of our proposed waypoint sampling method, we construct an area spanning $200\times200$ square meters from the KITTI odometry sequence-00. With waypoint sampling, we achieve a 2x speedup and reduced GPU memory consumption, all while maintaining the same reconstruction quality, as detailed in Table~\ref{table:waypoint_sampling}. Additionally, we reconstruct the entire area using poses derived from ORB-SLAM2, as visualized in Fig.~\ref{fig:structure}. This reconstruction of the entire road surface (covering $600\times600$ square meters) was completed in just two hours.

\subsection{Performance Comparision}
\label{s:exp:com}
\begin{figure*}[t!]
   \centering
   \includegraphics[width=1.0\linewidth]{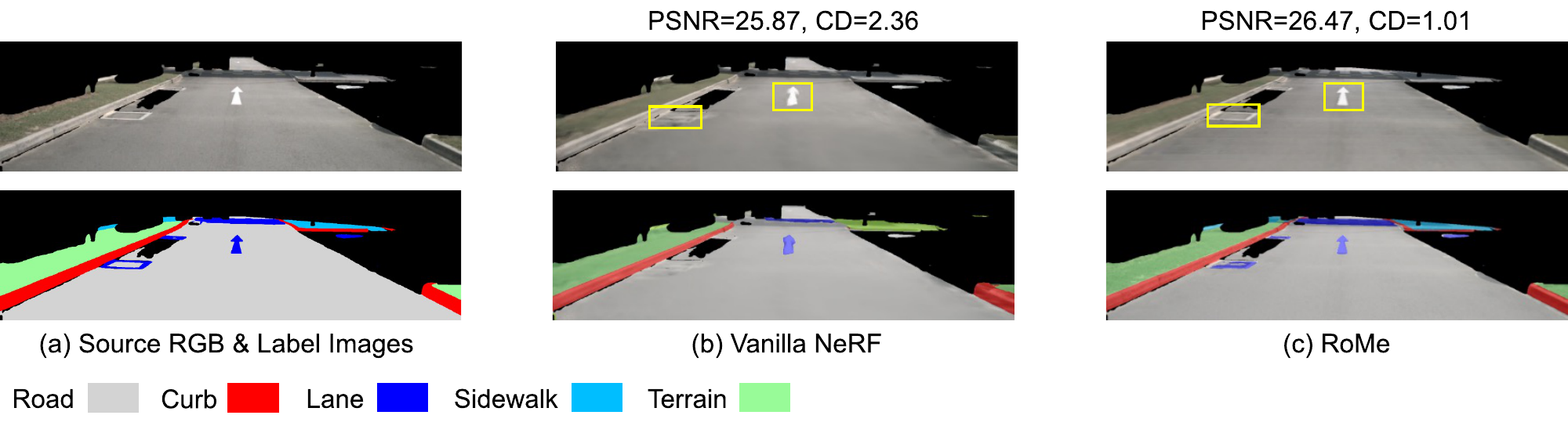}
   \caption{\small RGB and semantic reconstruction comparison. A segment from the nuScenes dataset is chosen, with three frames set aside for testing. The rest serve as training data. The yellow boxes highlight that RoMe captures finer details than the vanilla NeRF.}
   \label{fig:quantitative}
   \vspace{-1em}
\end{figure*}
To strike a balance between training speed and reconstruction quality, we conduct experiments on BEV resolution using the $scene$-$0391$ from the nuScenes dataset. The results are presented in Fig.~\ref{fig:rome_bevres}. A BEV resolution greater than or equal to 0.2m/pixel led to blurry reconstructions. Conversely, resolutions less than or equal to 0.05m/pixel added unnecessary computational overhead. Thus, a resolution of 0.1m/pixel (highlighted with a star) provided the optimal trade-off between quality and speed.

\begin{figure*}[t!]
   \centering
   \includegraphics[width=1\linewidth]{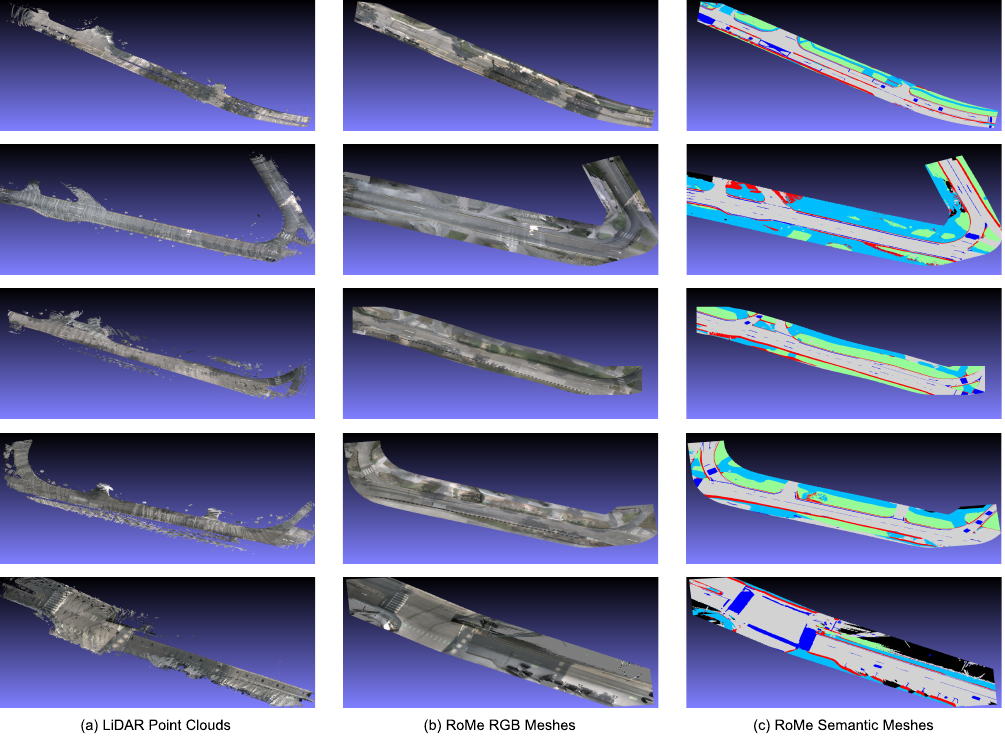}
   \caption{\small Visualization of LiDAR point clouds and RoMe meshes.}
   \label{fig:vis_mesh}
   \vspace{-1em}
\end{figure*}

\begin{figure*}[t!]
\renewcommand{\figurename}{Figure}
  \centering
  \includegraphics[width=0.8\linewidth]{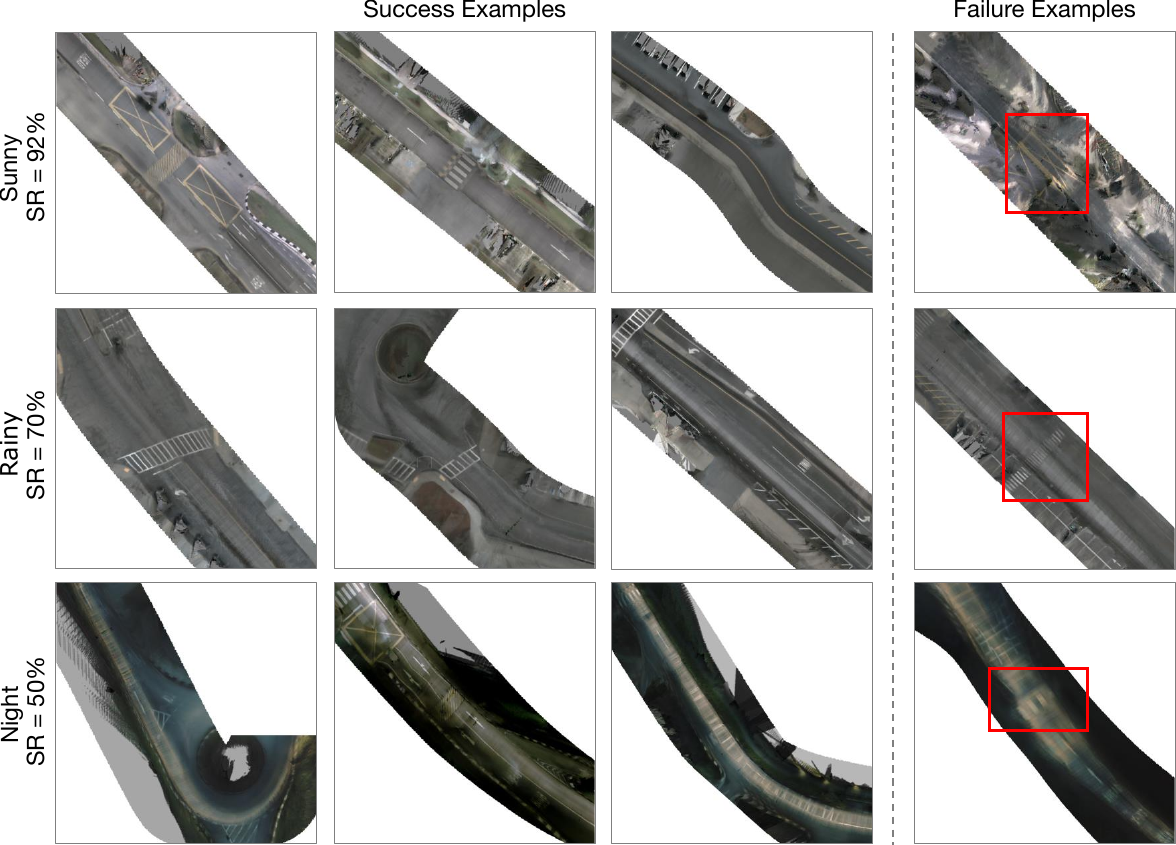}
  \caption{\small Evaluation of robustness. Our proposed RoMe achieves 92\%, 70\% and 50\% SR in sunny, rainy and night scenes, respectively. Blurry areas are marked in red boxes. SR denotes the Success Rate of each lighting condition.}
  \label{fig:robust}
\end{figure*}

\subsubsection{Comparison with COLMAP}
For comparison, we select the $scene$-$0655$ from the nuScenes dataset and masked all mobile obstacles. As presented in Fig.~\ref{fig:colmap}, the robustness of RoMe on moving objects significantly surpasses the COLMAP~\cite{schoenberger2016sfm}. The BEV mesh generated by COLMAP (with the Poisson mesher) tends to produce holes when encountering moving objects. In contrast, RoMe consistently generates a complete road mesh, provided at least one frame has a clear view of the road surface. Additionally, RoMe can simultaneously produce BEV semantics.
\begin{figure}[t!]
   \centering
   \includegraphics[width=1.0\linewidth]{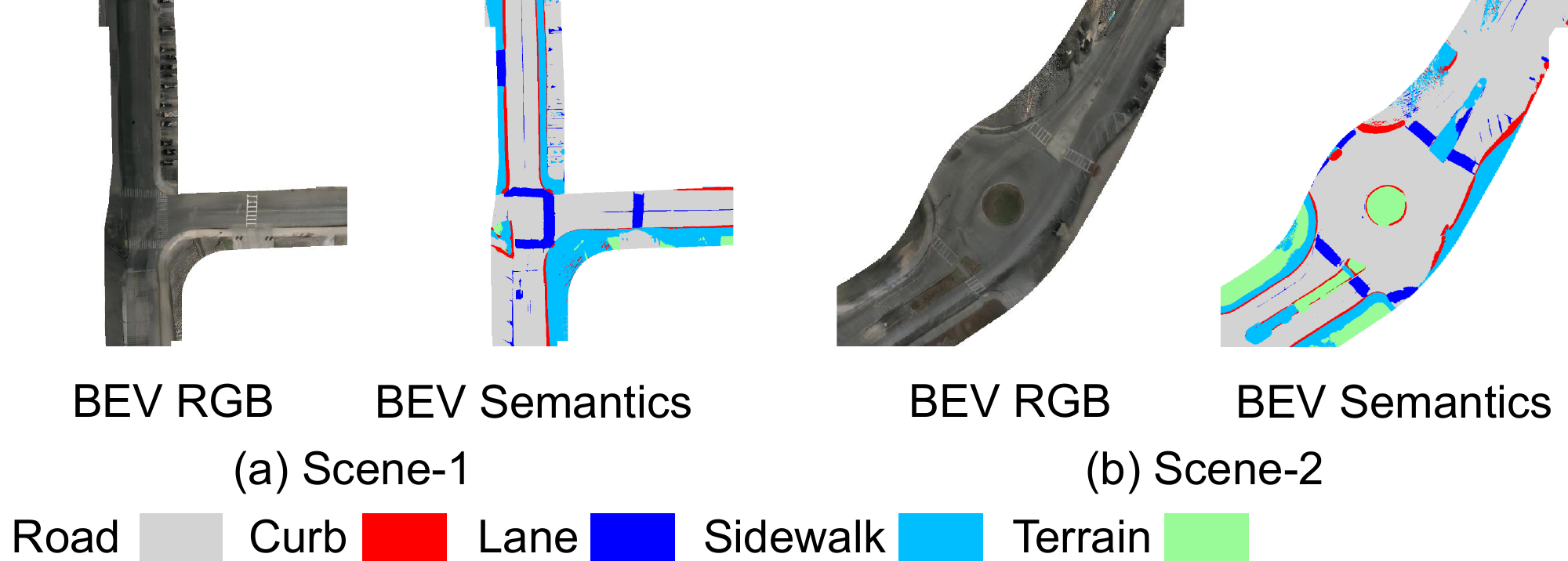}
   \caption{\small Results from the nuScenes dataset. The reconstructed road surface consistently represents only the immovable objects.}
   \label{fig:multi_scenes}
   \vspace{-2em}
\end{figure}
\subsubsection{Comparison with NeRF}
We sought to compare the capabilities of our proposed RoMe with the vanilla NeRF. For this purpose, we select a short clip ($scene$-$0990$) from the nuScenes dataset, ensuring it includes non-key frames to achieve higher image frame rates. Only images from the front camera are utilized. Fig.~\ref{fig:quantitative} showcases the RGB reconstruction alongside the segmentation results. The first column displays the source RGB and label images. The second column presents RGB images reconstructed by the vanilla NeRF and semantics segmented by Mask2Former~\cite{cheng2022masked}. The third column features RGB images and semantics reconstructed using RoMe. Our method delivers more realistic RGB reconstructions and precise semantic results. The road elements, highlighted in the yellow boxes, are more distinct than those in the vanilla NeRF. In a region spanning $70*70$ square meters, our method converged in approximately 8 minutes, whereas the vanilla NeRF required 20 hours. The original NeRF, due to its design, needs to restore depth across a broad range (e.g., $0\sim100$ meters) without depth supervision. In contrast, RoMe focuses on restoring elevations of less than 1 meter, which is more straightforward to optimize. The mesh representation inherently captures road surface features, which are predominantly flat but can exhibit significant changes at boundaries like curbs and slope edges.
Here, we verify the speed and accuracy of our proposed RoMe by quantitative comparison with NeRF:
\begin{outline}
 \1 \textbf{Speed}: We conduct experiments on various BEV ranges: from $100\times100$ to $300\times300$ square meters. Table~\ref{table:training_speed} details the training time comparison between F2-NeRF and our proposed RoMe. For small to medium scales, like $100\times100$ and $200\times200$ square meters, our proposed method has 2x to 4x speed up against F2-NeRF. We achieve a similar training speed for larger BEV ranges like $300\times300$ square meters while maintaining smaller memory footprints. The F2-NeRF fails because of out-of-memory (OOM) in large-scale reconstruction, especially with thousands of images.
 \1 \textbf{Accuracy}: We randomly select 100 scenes from the nuScenes dataset to evaluate the accuracy. Table~\ref{table:accuracy} details the reconstruction quality and accuracy comparisons between F2-NeRF and our proposed RoMe in terms of PSNR, mIoU, CD, and RMSE. Our method delivers higher accuracy in all metrics.
\end{outline}
 
It is worth noting that rather than a general reconstruction approach, RoMe is specially designed for road surface reconstruction to tackle the challenge of auto labelling in intelligent driving. In this scenario, NeRF has limited performance and robustness.
\begin{table}[t!]
\small
\centering
\renewcommand{\tablename}{TABLE}
   \caption{\small Training speed comparison. $100m\times100m$ means the total reconstruction areas are $100\times100$ square meters. Our proposed RoMe achieves faster speed and smaller memory footprints.}
   \begin{tabular}{c|c|c|c|c}
   \hline
    BEV range & Site ID & Image \# & F2-NeRF & RoMe (Ours)\\
   \hline
    \multirow{5}{*}{$100m\times100m$}
     & {0} & {1322} & {23.35min} & {6.28min} \\
     & {1} & {2306} & {30.25min} & {5.28min} \\
     & {2} & {1702} & {29.60min} & {7.89min} \\
     & {3} & {1694} & {27.93min} & {4.05min} \\
     & {4} & {1668} & {28.13min} & {6.98min} \\
    \hline
    \multirow{5}{*}{$200m\times200m$}
     & {5} & {2582} & {27.43min} & {15.16min} \\
     & {6} & {4505} & {37.32min} & {15.01min} \\
     & {7} & {3365} & {36.23min} & {8.30min} \\
     & {8} & {5535} & {OOM} & {9.37min} \\
     & {9} & {2860} & {30.57min} & {16.28min} \\
    \hline
    \multirow{5}{*}{$300m\times300m$}
     & {10} & {3612} & {35.08min} & {35.44min} \\
     & {11} & {4702} & {41.33min} & {41.94min} \\
     & {12} & {4800} & {OOM} & {18.94min} \\
     & {13} & {6071} & {OOM} & {46.95min} \\
     & {14} & {3350} & {33.92min} & {35.51min} \\
    \hline
   \end{tabular}
   \label{table:training_speed}
   \vspace{-1.8em}
\end{table}
\subsection{Robustness Validation}
\label{s:exp:robust}
\subsubsection{Multiple Scenes Validation}
We assess the robustness of RoMe by conducting experiments on 100 scenes selected from the NuScenes dataset. Specifically, we chose scenes characterized by favourable daytime weather conditions and trajectories spanning more than 100 meters. The experiments on these 100 scenes mirrored the approach described in Table~\ref{table:ablation_study}. We utilize all images for reconstruction and evaluated every image within each scene. Additionally, we fine-tuned the learning rate and adjusted the rotation and translation ranges for optimizing extrinsic. The results are summarized in Table~\ref{table:multi_scenes_ablation_study}.
\begin{figure}[t!]
   \begin{center}
      \includegraphics[width=1.0\linewidth]{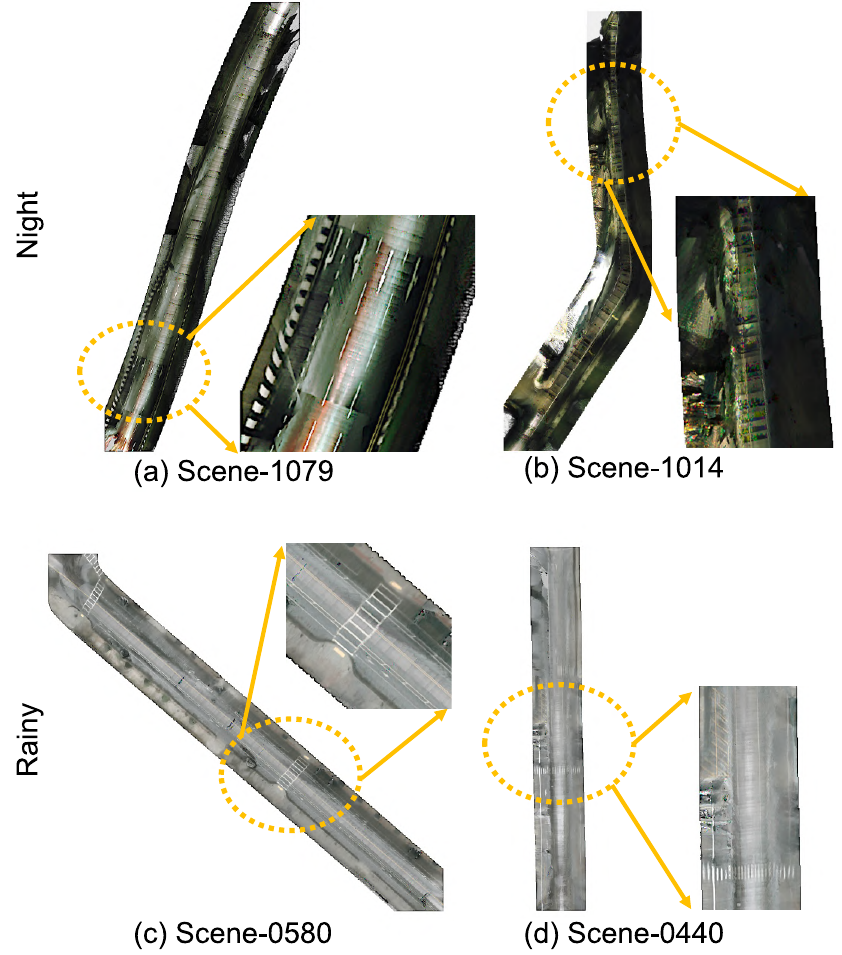}
   \end{center}
   \caption{\small Road surface reconstructions with RoMe during nighttime and rainy conditions. Exposures are slightly adjusted for a better view.}
   \label{fig:rome_night}
   \vspace{-1.0em}
\end{figure}
\begin{figure}[t!]
   \begin{center}
      \includegraphics[width=1.0\linewidth]{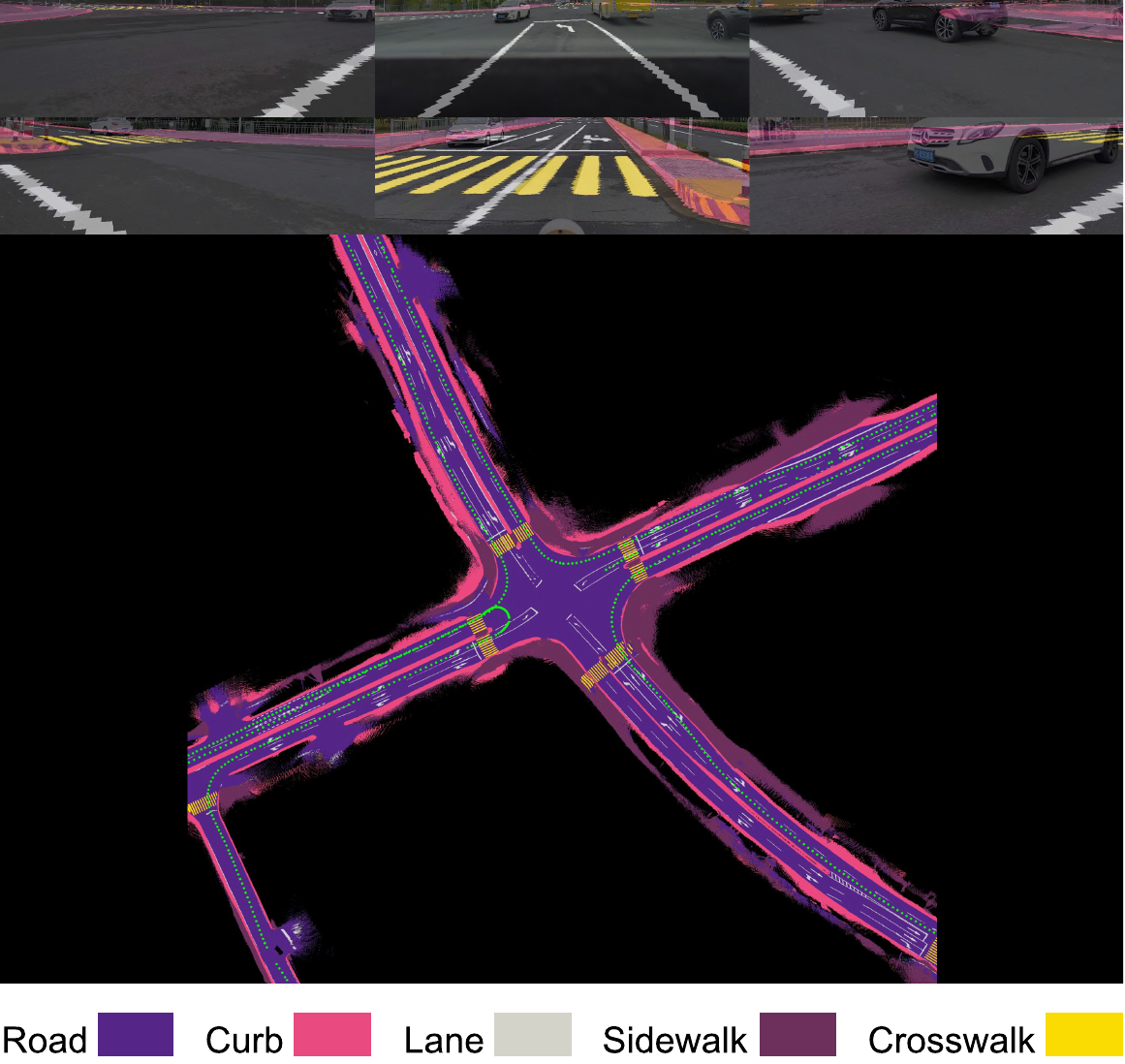}
   \end{center}
   \caption{\small Reconstruction visualization on wild data. The top two rows blend source RGB images with rendered semantics from the reconstructed road mesh. The bottom displays the reconstructed BEV semantics over a $300*300$ square meter area, with green dots indicating trajectories. The alignment between rendered semantics and source RGB images is precise, thanks to accurate poses, BEV semantics, and elevation reconstruction.}
   \label{fig:6v}
   \vspace{-1.0em}
\end{figure}
It's worth noting that while a high PSNR indicates good reconstruction quality for RGB images, it doesn't directly reflect the accuracy of the 3D structure reconstruction. This discrepancy arises because networks might overfit to RGB images rather than learning an accurate 3D structure, especially when there's insufficient posed image data. This phenomenon is evident in the third row of Table~\ref{table:multi_scenes_ablation_study}. By using a pose network to refine extrinsic and setting a larger learning rate and rotation/translation ranges, we achieve a higher PSNR but at the cost of an inaccurate 3D structure. This misalignment also affects the mIoU metric, as an incorrect 3D structure disrupts the alignment between source images and rendered semantics.

Fig.~\ref{fig:vis_mesh} offers a qualitative comparison between RoMe meshes and LiDAR point clouds. For clarity, we visualize the road points filtered by Mask2Former~\cite{cheng2022masked}. The RoMe meshes appear smooth and detailed, provide semantic information, and ultimately simplify the labelling process. In addition, we randomly and proportionally (based on original scene numbers) select 100 sunny, 20 rainy, and 20 night scenes to evaluate robustness for experiments. Success Rate (SR) and visual examples are detailed in Figure~\ref{fig:robust}. Scenes with unclear imagery and misalignment of road signs are considered failures. In rainy and night scenes, the road surface is unclear. Consequently, the reconstructed BEV RGBs are blurry and lack clarity.

\begin{table}[t!]
\renewcommand{\tablename}{TABLE}
\small
\centering
   \caption{\small Accuracy comparison between F2-NeRF and RoMe.}
   \begin{tabular}{c|c c c c}
   \hline
   methods & PSNR $\uparrow $ & mIoU ($\%$) $\uparrow $ & CD $\downarrow $ & RMSE$\downarrow $ \\
   \hline
   F2-NeRF & 9.19 & 31.45 & 15.93 & 6.98 \\
   RoMe (Ours) & \textbf{24.19} & \textbf{69.23} & \textbf{1.33} & \textbf{1.03} \\
   \hline
   \end{tabular}
   \label{table:accuracy}
   \vspace{-1.5em}
\end{table}
\subsubsection{Multiple Scenes Merging}
RoMe can seamlessly integrate different scenes as long as they share common positions. Fig.~\ref{fig:multi_scenes} showcases the results of merging multiple scenes. Both $Scene$-$1$ and $Scene$-$2$ are composites of four individual scenes. The smooth transitions between scenes are attributed to the precise camera poses and the ability to optimize the extrinsic. However, when there are significant differences in weather conditions, we prioritize semantics reconstruction, as semantics remain consistent across varying lighting conditions, unlike RGB images.
\begin{table*}[t!]
\small
\centering
   \setlength{\tabcolsep}{10pt}
   \caption{\small Multiple scenes ablation study. The table details the impact of optimizing elevation and extrinsic on reconstruction quality across various metrics. The unit of rotation is in degrees and translation in meters. A larger extrinsic learning rate and rotation/translation ranges can lead to image overfitting, resulting in a higher PSNR but an inaccurate 3D structure.}
   \begin{tabular}{c c c c c|c c c c}
   \hline
    Opt. elevation & Opt. extrinsic & extrinsic LR & rotation & translation  & PSNR $\uparrow $ & mIoU ($\%$) $\uparrow $ & CD $\downarrow $ & RMSE$\downarrow $\\
   \hline
    $\times$ & $\times$ & $- $ & $- $ & $- $ & 23.90 & 67.00 & 1.25 & 1.09\\
    $\surd$ & $\times$ & $- $ & $- $ & $- $ & 23.93 & 67.16 & 1.24 & 1.09\\
    $\surd$ & $\surd$ & 0.01 & 0.5 & 0.5 & \textbf{24.36} & 64.86 & 5.77 & 1.36\\
    $\surd$ & $\surd$ & 0.002 & 0.1 & 0.1 & 24.19 & \textbf{69.23} & \textbf{1.13} & \textbf{1.03}\\
   \hline
   \end{tabular}
   \label{table:multi_scenes_ablation_study}
\end{table*}
\subsection{Limitations and Applications}
\noindent{\bf Limitations:}
RoMe can reconstruct road surfaces on rainy days or at night, as shown in Fig.~\ref{fig:rome_night}: (a) and (c) are promising since light conditions are tolerable. When encountering worse light conditions, it is challenging to reconstruct the road surface, as shown in (b) and (d). While RoMe demonstrates the capability to reconstruct road surfaces under various conditions, including rainy days or nighttime scenarios, its performance can vary based on the severity of environmental conditions. Thus, the reconstruction quality can degrade in more adverse lighting conditions. This limitation underscores the need for further enhancements to RoMe's adaptability in diverse and challenging scenarios. Moreover, it should be emphasized that our proposed RoMe is designed for road surface reconstruction and unsuitable for novel view synthesis tasks (aka. general reconstruction from NeRF).

After the open-source release of the RoMe at GitHub$^{1}$, there have been multiple concurrent works~\cite{wu2024emie,feng2024rogs}, that have improved upon it. We hope that RoMe and the concurrent works will provide more inspiration to followers.
\let\thefootnote\relax\footnotetext{$^{1}$ \url{Github: https://github.com/DRosemei/RoMe}}

\noindent{\bf Applications:}
We applied RoMe to wild data, showcasing its versatility. Fig.~\ref{fig:6v} presents our reconstruction of an intersection. The alignment between the rendered semantics and the source RGB images is evident, demonstrating the precision of our method. This precision facilitates easy annotation of BEV lanes, curbs, arrows, crosswalks, and other static road elements directly on the road mesh.

\begin{figure}[t!]
   \begin{center}
      \includegraphics[width=1.0\linewidth]{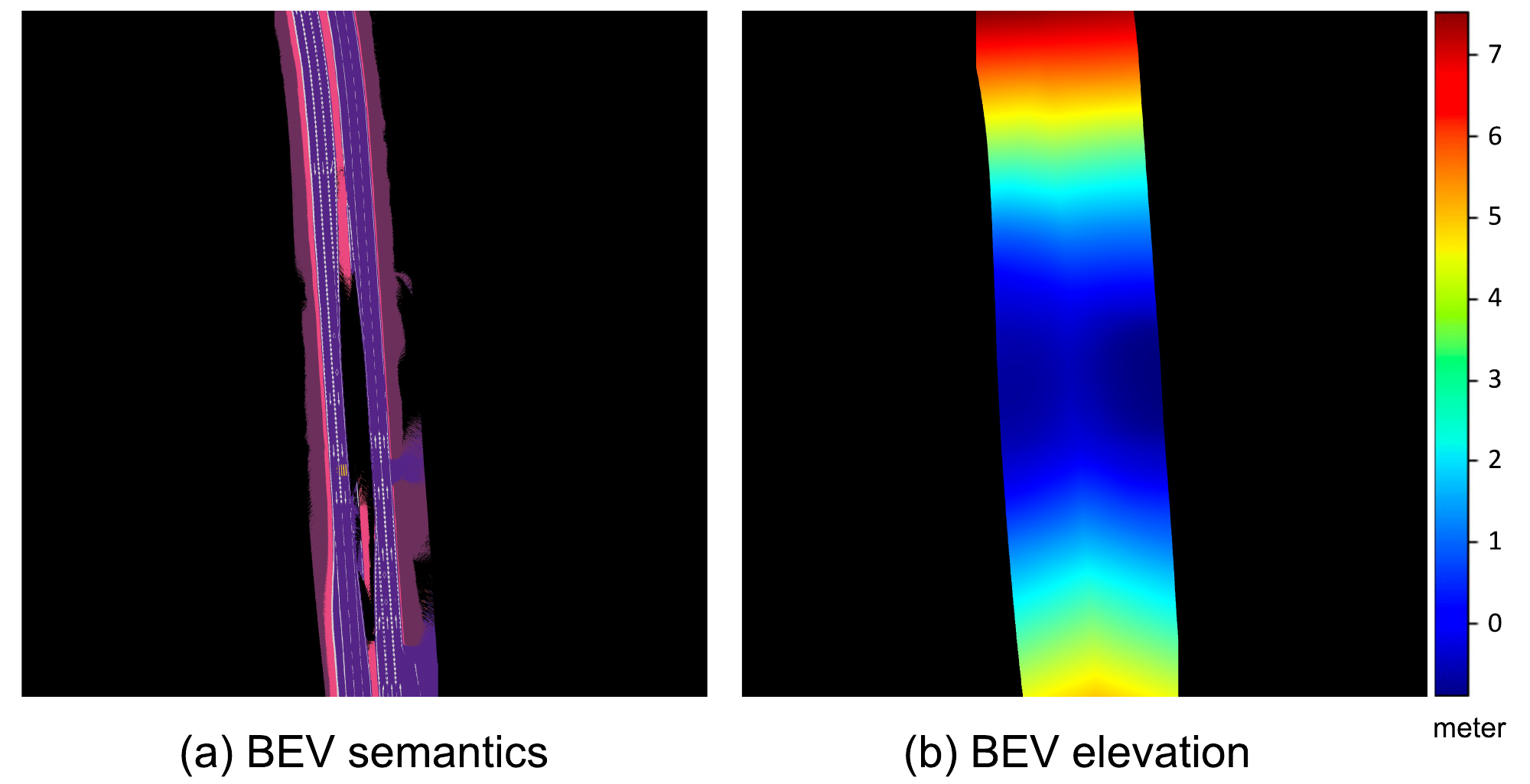}
   \end{center}
   \caption{\small Visualization of the reconstructed steep slope. The left side depicts BEV semantics, while the right side illustrates BEV elevation, ranging from -0.8 meters to over 7 meters. RoMe's capability to accurately reconstruct such varied elevations is evident.}
   \label{fig:chongqing}
\end{figure}

\begin{figure}[t!]
   \begin{center}
      \includegraphics[width=1.0\linewidth]{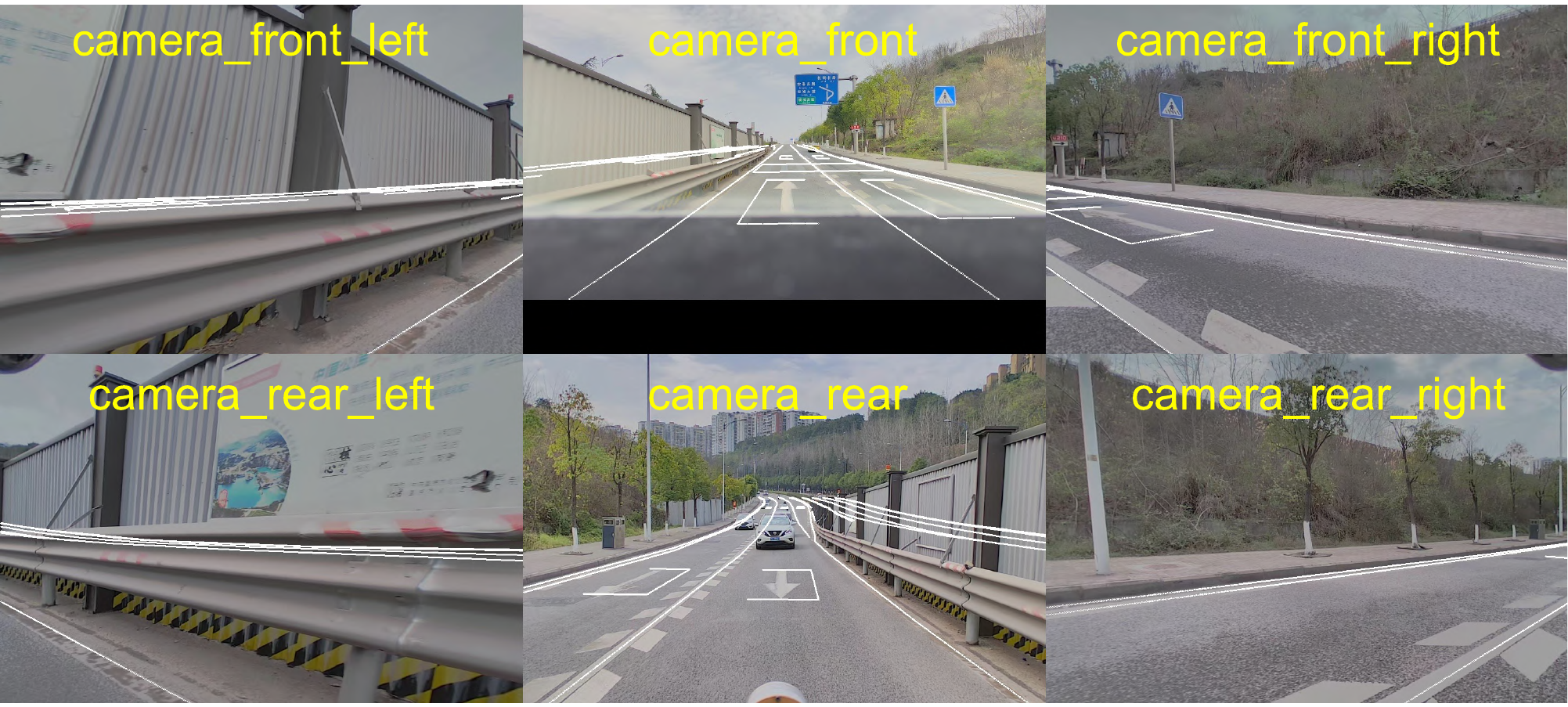}
   \end{center}
   \caption{\small Reprojection visualization on the reconstructed steep slope. Manually labelled lanes and arrows align seamlessly with road signs and lanes in the source images, validating the accuracy of our 3D road surface reconstruction.}
   \label{fig:chongqing_6v}
   \vspace{-1.5em}
\end{figure}

To further illustrate the strength of learning BEV elevation, we selected a scene in the city of Chongqing (China) characterized by a steep slope. In addition to the method above, we utilize Structure from Motion (SfM) or MVS points generated by COLMAP for precise supervision. Fig.~\ref{fig:chongqing} displays our reconstruction. The left side represents BEV semantics, while the right side showcases BEV elevation, which varies from -0.8 meters to over 7 meters. Despite the significant elevation changes, RoMe provides a clear and accurate reconstruction. This accuracy is further demonstrated in Fig.~\ref{fig:chongqing_6v}, where manually labelled lanes and arrows align perfectly with road signs and lanes, even over an elevation range exceeding 8 meters.

%% file: conclusion.tex
\section{CONCLUSION}
\label{s:conclusion}
Throughout this study, we have delved into the intricacies of road surface reconstruction, introducing RoMe as a groundbreaking solution tailored for expansive environments. RoMe stands out due to its unique approach, leveraging a mesh representation that ensures a robust reconstruction of road surfaces, seamlessly aligning with semantic data. This alignment is pivotal, especially when considering the challenges of large-scale reconstructions, where even minor misalignments can lead to significant inaccuracies.

Our evaluations, spanning areas as vast as $600*600$ square meters and encompassing renowned datasets like nuScenes and KITTI, have consistently showcased RoMe's superiority in terms of accuracy, speed, and resilience, particularly when compared to existing methods like the vanilla NeRF. The waypoint sampling strategy, a hallmark of RoMe, not only accelerates the training process but also optimizes computational resources. By reconstructing road surfaces in segmented regions and then integrating them during training, RoMe demonstrates its adaptability to large-scale environments without compromising on precision.

Moreover, introducing the extrinsic optimization module addresses a critical challenge in road surface reconstruction: the potential inaccuracies stemming from extrinsic calibration. This module, combined with RoMe's inherent design, ensures that the framework remains robust even in diverse and challenging scenarios, as evidenced by our experiments on both public and wild data.

In the context of autonomous driving, precision is of utmost importance. RoMe emerges as a transformative solution. Its ability to provide accurate reconstructions paves the way for automating the labeling process, a crucial step toward the realization of fully autonomous vehicles. As we move forward, the innovations presented in this study underscore the potential of RoMe to revolutionize road surface reconstruction and its broader applications in autonomous driving.